\def\year{2019}\relax
\documentclass[letterpaper]{article} 
\usepackage{aaai19}  
\usepackage{times}  
\usepackage{helvet}  
\usepackage{courier}  
\usepackage{url}  
\usepackage{graphicx}  
\usepackage{rotating} 
\usepackage{booktabs} 
\usepackage{multirow}
\usepackage{rotating}
\usepackage{amsfonts} 
\usepackage{xcolor}

\makeatletter
\g@addto@macro{\UrlBreaks}{\do\/\do\-\do_}
\makeatother
\usepackage{amssymb}
\usepackage{amsmath}
\usepackage{caption}

\usepackage{color, colortbl}
\definecolor{Cyan}{rgb}{0.88,1,1}
\definecolor{Pink}{rgb}{1,0.75,0.80}
\newcommand{\cellcolorandbold}[2]{\cellcolor{#1}{\textbf{#2}}}

\begin{document}
\title{Investigating Human + Machine Complementarity for Recidivism Predictions}

\author{Sarah Tan \\
Cornell University \\
\And 
Julius Adebayo\\
MIT \\
\And 
Kori Inkpen \\
Microsoft Research\\
\And 
Ece Kamar \\
Microsoft Research \\
}
\maketitle

\begin{abstract}
When might human input help (or not) when assessing risk in fairness domains? 
Dressel and Farid (\citeyear{dressel2018accuracy}) asked Mechanical Turk workers to evaluate a subset of defendants in the ProPublica COMPAS data for risk of recidivism, and concluded that COMPAS predictions were no more accurate or fair than predictions made by humans. We delve deeper into this claim to explore differences in human and algorithmic decision making. We construct a Human Risk Score based on the predictions made by multiple Turk workers, characterize the features that determine agreement and disagreement between COMPAS and Human Scores, and construct hybrid Human+Machine models to predict recidivism. Our key finding is that on this data set, Human and COMPAS decision making differed, but not in ways that could be leveraged to significantly improve ground-truth prediction. We present the results of our analyses and suggestions for data collection best practices to leverage complementary strengths of human and machines in the fairness domain.
\end{abstract}

\section{Introduction}

The criminal justice field has used forecasting tools to perform risk assessment since the 1920s
(see \citeauthor{Gendreau1996meta} \shortcite{Gendreau1996meta} and \citeauthor{Andrews2006recent} \shortcite{Andrews2006recent} for  meta-reviews). More recently, machine learning approaches are being used to inform bail and sentencing decisions \cite{berk2013statistical,Angwin2016machinebias}. But, this raises concerns around accuracy, fairness, and transparency of risk assessment systems \cite{Andrews2006recent,drake2014predicting,zeng2017interpretable}.  

One ongoing debate is whether risk assessment systems are superior to human judgment. \citeauthor{Grove2000clinical} \shortcite{Grove2000clinical} conducted a meta-analysis of 136 studies of human health and behavior to assess clinical (Human) versus mechanical (Machine) predictions. Their results revealed that on average, machine predictions were 10\% more accurate than Human predictions, however there were some studies that showed no improvements and even a few cases where human predictions were more accurate. All three recidivism studies in  \citeauthor{Grove2000clinical}'s analysis that tracked accuracy revealed similar levels of accuracy between human and machine predictions. On the other hand, \citeauthor{kleinberg2017human} \shortcite{kleinberg2017human} found that expert Humans (judges)'s decisions can sometimes be highly variable and biased by unobserved, irrelevant features.

In a recent study related to Human vs. Machine predictions of recidivism, \citeauthor{dressel2018accuracy} \shortcite{dressel2018accuracy} showed that a widely used commercial risk assessment system for recidivism -- COMPAS -- was no more accurate or fair than predictions made by people with little to no experience in criminal justice. They sampled 1,000 defendants from the ProPublica COMPAS data \cite{propublica2018compas} and asked Mechanical Turk workers to predict whether a defendant would recidivate within two years (the same label predicted by COMPAS).  They also ran a second variant of their study where defendants' race was revealed. They did not find Human and COMPAS accuracies to be significantly different (COMPAS: 65.2\%, Humans without defendant race information: 67.0\%, and Humans with race information: 66.5\%).

Although the \citeauthor{dressel2018accuracy} study demonstrated that COMPAS and Human predictions were comparable, it was unclear whether COMPAS and Humans were accurate on the same or disjoint sets of defendants. Significant overlap would suggest that the Humans and COMPAS make similar assessments; less overlap suggests that human reasoning differed from machine analysis. Humans may have access to additional information or context not available to algorithmic systems; machines may not be influenced by the same biases that plague human judgment. Instead of focusing on the superiority (or lack thereof) of algorithmic systems compared to human judgment, we explore the similarities and differences between Human and COMPAS decisions to determine whether a hybrid approach that combines the strengths and addressess the weaknesses of human and machine decision making is possible. 

\begin{table*}
\centering
\footnotesize
  \caption{Characterizing agreement and disagreement between COMPAS decisions, Human decisions, and ground truth. The number of defendants and characteristics for each of the eight cases are described.}
  \label{tab:cases}
  \begin{tabular}{ccccccccc}
    \toprule
    Case & COMPAS & Human & Ground & Agreement & Correctness & \% Defendants  & Feature \\
    & Score & Score & Truth & & &  & Characteristics* \\
    \toprule
    1 & High & High & Yes & Agree & Both correct & \multirow{2}{*}{49.0\%} & 1.5 $<$ Priors $\leq$ 12.5\\
    2 & Low & Low & No & Agree & Both correct &  & 23.5 $<$ Age $\leq$ 48.5 \& Priors $<$ 1.5 \\
    \midrule
    3 & High & Low & Yes & Disagree & COMPAS correct & \multirow{2}{*}{16.2\%} & 23.5 $<$ Age $\leq$ 48.5  \& Priors $<$ 0.5 \\
    4 & Low & High & No & Disagree & COMPAS correct &  & 1.5 $<$ Priors $\leq$ 5.5 \& Age $>$32 \\
    5 & Low & High & Yes & Disagree & Human correct & \multirow{2}{*}{15.9\%} & Similar to Case 4 \\
    6 & High & Low & No & Disagree & Human correct &  & Similar to Case 3 \\
    \midrule
     7 & High & High & No & Agree & Both incorrect & \multirow{2}{*}{18.9\%} & \multirow{2}{*}{No pattern, similar to Cases 1-6} \\
    8 & Low & Low & Yes & Agree & Both incorrect  & &  \\
  \bottomrule
 \multicolumn{9}{l}{* Characteristics determined by decision tree (Figure \ref{fig:tree8}) and clustering analysis. See more details in Analysis and Results section. }
\end{tabular}
\end{table*}

Our contributions in this paper are: \begin{itemize}
    \item We analyze, on recidivism data, how human and machine decisions differ and how they make errors. 
    \item We characterize areas of agreement and disagreement between human and machine decision making to better understand their complementarity.
\item We investigate if hybrid models can leverage differences in human and machine decision making to improve recidivism prediction. \item Based on our findings, we discuss shortcomings of existing data sets and make recommendations for data collection best practices for future study of hybrid decision making in the fairness domain.
\end{itemize}

\section{Related Work}
\textbf{Humans and decision making.} In addition to the work mentioned in the Introduction, Lakkaraju et. al., (\citeyear{lakkaraju2017selective}) showed that analyses of recidivism based on human decisions are further complicated by the ``selective labels'' problem, where observability of outcomes are affected by judges' release decisions. Other work studied how humans perceive different features as fair or not \cite{grgic2018human}.

\textbf{Hybrid models.} Investigations across different domains identify that humans and machines have weaknesses and complementary abilities, thus suggesting benefits from hybrid models. In medicine, recent research showed that existing machine learning models with lower accuracy rates than human experts can decrease expert error rates by 85\% \cite{wang2016DeepLF}. On challenging face recognition tasks, combining multiple expert opinions does not improve task accuracy, however complementing an expert with a inferior face recognition system can \cite{phillips2018face}. On the other hand, research on complementary computing demonstrated how humans and machines can be more effective together in problem solving \cite{horvitz2007complementary} and image classification tasks \cite{kamar2012combining}. 

\textbf{Diagnosing errors.}
The key to aggregating machine and human analyses to improve performance is understanding where and how machines and humans fail. Various approaches have been proposed for understanding where machine errors come from. \citeauthor{lakkaraju2017identifying} \shortcite{lakkaraju2017identifying} defined \textit{unknown unknowns} as cases where the model is highly confident of its predictions but is wrong.  
\citeauthor{kulesza2015principles} \shortcite{kulesza2015principles} uses human input to interactively correct a model. Another approach is to distill black-box model decisions to interpretable model classes to explain model failures \cite{nushi2018accountable,tan2017detecting}. 
We follow a similar approach of utilizing interpretable models to analyze how machines and humans reason about recidivism, when and how their decisions differ and how they can be aggregated.  

\section{Approach} 

\subsection{Constructing Human risk score} Our goal in this paper is to compare COMPAS and Human decision making. Before doing so, one may ask if the decisions made by Mechanical Turk workers in this data are \emph{internally} consistent, or, in other words, if different Turk workers assess risk similarly for the same defendant. Large agreement among Turk workers increases confidence that our subsequent findings based on generating Human scores from Turk worker predictions generalize to Human decision making. 
We find that on average, 80\% of the 20 Turk workers that assess the same defendant agree with each other. This is a high level of agreement, particularly for Mechanical Turk, where spam labeling is commonly observed \cite{ipeirotis2010quality}. 
Hence, we perform a majority aggregation of Turk workers' predictions to assemble a Human risk score for recidivism risk, $h_j$. Specifically, we construct $h_j$ by taking the mean prediction across 20 Turk workers for each defendant: let $h_{ij}$ be Turk worker $i$'s prediction for defendant $j$ where $h_{ij} \in \{0,1\}$, $i=1, \ldots, 20, j=1, \ldots, 1000$, we take $h_j = \sum_i h_{ij} / 2$, dividing by two to scale $h_j$ to 1-10, which is COMPAS' scale. We constructed scores for both conditions mentioned in the Introduction - a with-race Human score (HWR) for when Turk workers were told the defendants' race, and a no-race version (HNR). 

For each score, we find the optimal cutoff point to binarize the score by computing calibration, false positive, and false negative rates at various cutoff points from 1 to 10. COMPAS, HNR, and HWR scores have approximately equal accuracy, false positive, and false negative rates at the cutoff point of $>=5$ (Figure \ref{fig:calibration} in Appendix). Hence, we chose this cutoff point for all three scores. Note that Northpointe, the creator of COMPAS, also uses a $>=5$ cut-off \cite{blomberg2010validation}, and $>=5$ is implied by \citeauthor{dressel2018accuracy}'s use of a ``wisdom-of-the-crowd'' based majority rules criterion. 

\subsection{Partitioning by agreement and correctness}
\label{sec:partitioning}
We now sketch our approach towards studying how COMPAS and Human scores agree or disagree, and interact with ground truth. Table \ref{tab:cases} describes eight possible combinations of two binary risk scores and ground truth. These eight combinations can be grouped into the four partitions illustrated in Table \ref{tab:cases}: Both correct, Both incorrect, Human correct, and COMPAS correct.  

Comparing the level of agreement and correctness between the Human and COMPAS scores, we found that almost 50\% of the time, Humans and COMPAS agree and are correct (Table \ref{tab:cases}).  However, for the remaining 50\% of defendants, either one, or both scores were incorrect. This suggests that if error regions of COMPAS and Humans do not perfectly overlap \textbf{and can be characterized}, then decision-making processes can potentially be improved through utilizing the complementary views of humans and machines.

When both risk scores agree and are correct, either score will return the same prediction, hence it does not matter which is used (in terms of accuracy). The space where both scores agree but are incorrect according to ground truth is a blind spot for COMPAS and Humans, also called \textit{unknown unknowns} \cite{lakkaraju2017identifying}. To characterize the space of agreement or disagreement between COMPAS and Human scores, we use clustering and decision trees. Table \ref{tab:cases} summarizes our findings of the features that characterize each case. Finally, when COMPAS and Human scores disagree (Cases 3-6 in Table \ref{tab:cases}) we train hybrid risk scoring models to see if they can leverage disagreement between the two scores to improve on the accuracy of single scores. 

\subsection{Hybrid models}
\label{sec:hybridmodels}
The simplest hybrid model is an average of two risk scores. We train a slightly more sophisticated model - a \textbf{weighted average} hybrid model that learns the optimal linear combination of two risk scores to predict ground truth. We also train \textbf{direct} and \textbf{indirect} hybrid models: direct models directly predict ground truth recidivism as a function of defendant features and Turk worker features (race, gender, and age) and the two scores; indirect models first predict which of the two scores to pick, then take that score's prediction of ground truth.  We test the hybrid models against \textbf{random} and  \textbf{single} score baselines. We use two types of random baselines: random ground truth labels, and random risk score. Single score baselines are COMPAS or Human scores on their own (1-10 scale, or binarized at $>=$5), or models trained with defendant and Turk worker features and the single score to predict ground truth. All hybrid and single models in this paper were trained using the random forest model class, a model class shown to perform well on many problems \cite{caruana2006empirical}. 
We use area under the ROC curve (AUC) as our main accuracy measure, in line with several papers measuring recidivism \cite{ussc2004measuring}, but we also report other metrics in Appendix D. All metrics are reported over ten 80\%-20\% train-test splits to account for  variability between test sets.  See Appendix A for more details on the hybrid models and error metrics.

\section{Analysis and Results}
In this section, we report our findings of COMPAS and Human complementarity in terms of predictive performance and decision making, characterize the space of COMPAS and Human agreement and disagreement, and discuss results from our hybrid models.

\subsection{COMPAS vs. Humans: Predictive Performance}
Across 1,000 defendants, Human scores have slightly higher means than COMPAS (mean HNR 5.1, HWR 5.2, COMPAS 4.6, all on 1-10 scale), and the Human scores are more correlated with each other than with COMPAS (COMPAS and HNR correlation 0.52, COMPAS and HWR 0.53, HNR and HWR 0.93). 

Table \ref{tab:riskscores} displays the predictive performance of COMPAS and Human scores, on all defendants and by race  (this is similar to Table 1 in \citeauthor{dressel2018accuracy} \shortcite{dressel2018accuracy}). Most scores achieved accuracies around 0.66 and AUCs around 0.70 when evaluated on all races, blacks, or whites. All scores performed slightly worse for other races at approximately 0.65; there are only a small number of these defendants in the data (9\%). These findings replicate \citeauthor{dressel2018accuracy}'s findings  when evaluating accuracies of the three scores. 

Table \ref{tab:riskscores_recid} presents predictive performance separated by recidivism status: whether the defendant recidivated or not. Here, Humans were better at predicting defendants who recidivate, while COMPAS was better at predicting defendants who do not recidivate. In other words, on this data, Humans tended to have higher true positive rates (and hence lower false negative rates) and COMPAS tended to have  higher true negative rates (and hence lower false positive rates).

\begin{table}[ht]
\centering
\footnotesize
\caption{COMPAS and Human accuracy and AUC when predicting ground truth recidivism.}
\label{tab:riskscores}
\begin{tabular}{lllllll}
  \toprule
  & \multicolumn{3}{c}{Accuracy} & \multicolumn{3}{c}{AUC} \\
  \cmidrule(lr){2-4}\cmidrule(lr){5-7}
Race & C & HNR & HWR & C & HNR & HWR \\ 
  \midrule
All & 0.65 & \cellcolorandbold{Cyan}{0.66} & \cellcolorandbold{Cyan}{0.66} & 0.70 & 0.71 & 0.71 \\ 
  Black & \cellcolorandbold{Cyan}{0.68} & 0.66 & 0.65 & 0.70 & 0.69 & 0.69 \\ 
  White & \cellcolorandbold{Cyan}{0.66} & \cellcolorandbold{Cyan}{0.66} & 0.64 &  0.71 & 0.69 & 0.70 \\ 
    Other & 0.65 & 0.60 & \cellcolorandbold{Cyan}{0.66} & 0.64 & 0.65 & 0.65 \\ 
   \bottomrule
\end{tabular}
\end{table}

\begin{table}[ht]
\centering
\footnotesize
\caption{Refinement of Table \ref{tab:riskscores} by recidivism status. Left: defendants who \textbf{do} recidivate. Right: defendants who \textbf{do not} recidivate. Only accuracies are displayed because AUC cannot be calculated when ground truth only has one value (``yes'' for do recidivate, ``no'' for do not recidivate).}
\label{tab:riskscores_recid}
\begin{tabular}{lllllll}
  \toprule
  & \multicolumn{6}{c}{Accuracy} \\
  \cmidrule(lr){2-7}
  & \multicolumn{3}{c}{Do recidivate} & \multicolumn{3}{c}{Do not recidivate} \\
  \cmidrule(lr){2-4}\cmidrule(lr){5-7}
Race & C & HNR & HWR  & C & HNR & HWR\\ 
  \midrule
All & 0.62 & 0.68 & \cellcolorandbold{Cyan}{0.69} & \cellcolorandbold{Cyan}{0.69} & 0.64 & 0.63 \\ 
  Black & \cellcolorandbold{Cyan}{0.74} & \cellcolorandbold{Cyan}{0.74} & 0.70 & 0.61 & 0.55 & 0.59 \\ 
  White & 0.60 & 0.50 & 0.59& 0.69 & \cellcolorandbold{Cyan}{0.75} & 0.68 \\ 
    Other & 0.38 & 0.59 & 0.65 & \cellcolorandbold{Cyan}{0.80} & 0.61 & 0.66 \\ 
   \bottomrule
\end{tabular}
\end{table}

\begin{figure*}[htb]
\begin{center}
\begin{tabular}{cc}
\begin{turn}{90}\hspace{0.75cm} \scriptsize{Feature Importance}\end{turn}
\includegraphics[width=0.23\textwidth,height=1.4in]{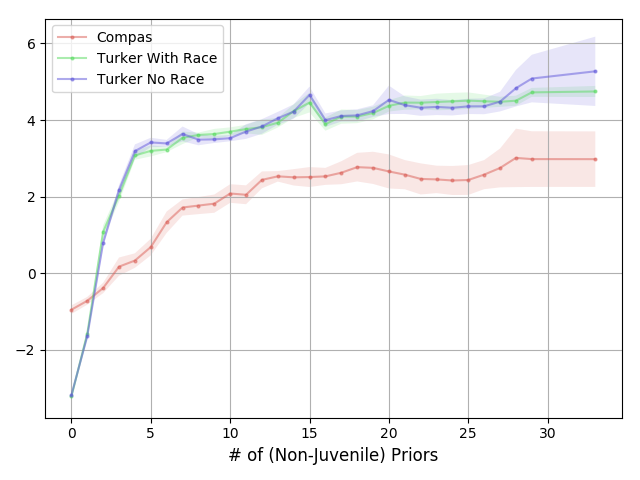} 
\includegraphics[width=0.23\textwidth,height=1.4in]{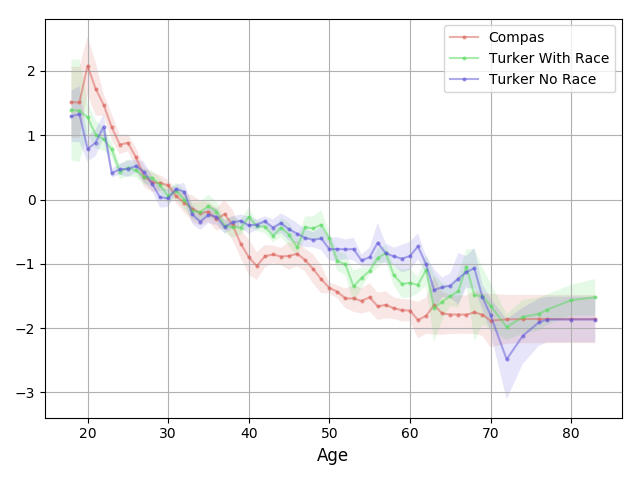} 
\includegraphics[width=0.23\textwidth,height=1.4in]{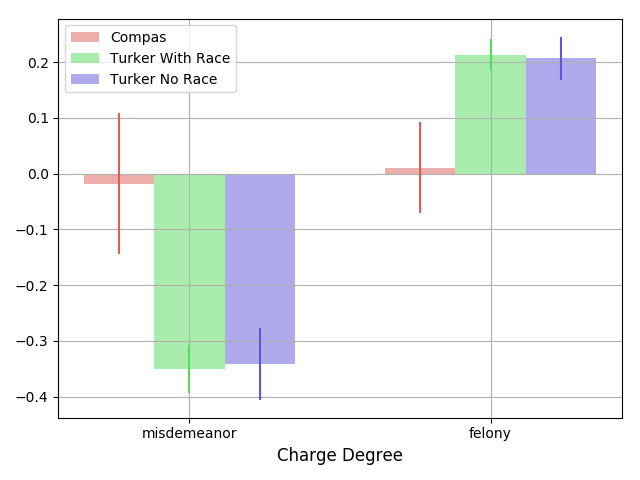} 
\includegraphics[width=0.23\textwidth,height=1.4in]{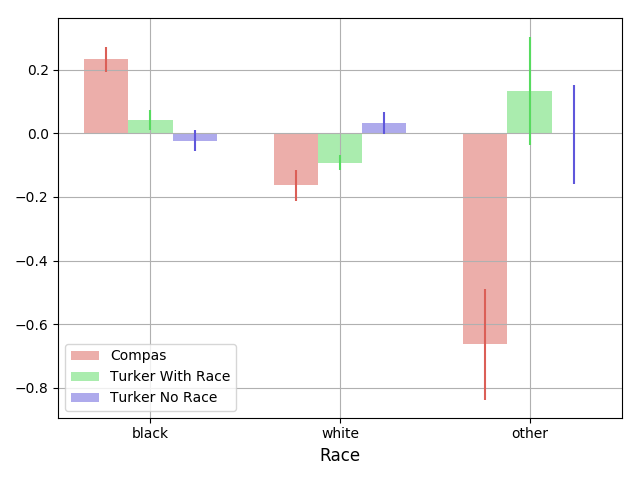}
\end{tabular}
\caption{Predicting COMPAS (red), HWR (green), and HNR (purple) scores from features. The larger the y-axis magnitude, the more important the feature. ``Number of priors'', with y-axis scale -3 to 6, is the most important feature for all three scores, followed by ``age''.}
\label{fig:gam}
\end{center}
\end{figure*}

We see similar effects for the level of agreement between risk scores, race, and ground truth. COMPAS and Humans demonstrate higher levels of agreement for correctly predicting that black defendants will recidivate, but their level of agreement drops significantly for white or other race defendants who recidivate.  The opposite is true for defendants who do not recidivate. COMPAS and Humans have higher levels of agreement for correctly predicting that white and other race defendants will not recidivate, but this level of agreement drops for black defendants who do not recidivate.

\subsection{COMPAS vs. Humans: Decision Making} Which features are most important in COMPAS and Human decision making? It is known that COMPAS scores can be predicted from only a few features, in particular the ``number of priors'' and age \cite{Chouldechova2017Fairer,angelino2017learning,tan2017detecting}. To determine if Human decision making places more importance on other features, we trained interpretable models to predict each of the three scores. All three models saw the same set of features -- age, race, sex, number of juvenile misdemeanors, number of juvenile felonies, number of (non-juvenile) priors, crime charge degree (misdemeanor or felony), and crime charge. First, we trained iGAM models, a type of additive model based on  nonparametric base learners \cite{caruana2015intelligible}. Figure \ref{fig:gam} illustrates the importance of four of these features for predicting each score. \textbf{Like COMPAS, the two most important features in Human decision making are the ``number of priors'' and ``age''.} However, Human scores place more weight on the ``number of priors'' and ``charge degree'' features than COMPAS, whereas age's impact is similar for COMPAS and Human scores. Decision trees trained to predict each of the three risk scores confirm that ``number of priors'' is the most important feature, with every tree's root node splitting on this feature. 

Including  race  when  predicting these scores,  even when the scores may not have seen race, returns some interesting findings. Recall that HNR scores were generated from Turk workers who were not told the defendants' race. We considered the impact of race on Human recidivism predictions, by comparing the importance of the race feature on HWR (green) and HNR (purple) scores in Figure \ref{fig:gam}. We find that black defendants were assessed to have slightly higher recidivism risk by Turk workers when told of their race. The decision tree predicting the difference of HWR and HNR scores in  Figure \ref{fig:tree3} also agreed with this finding, returning a first split on race where white defendants were assigned slightly lower risk (-0.16) in the Human with-race condition, and black and other race individuals were assigned slightly higher risk (+0.14). In contrast, both decision trees predicting the difference between COMPAS and HWR scores, as well as COMPAS and HNR scores, split on ``number of priors'' and age but not race. 

Hence, even though revealing race did not significantly affect the predictive performance of Humans for ground truth, as found by \citeauthor{dressel2018accuracy}, including race appeared to have  slightly affected Humans' perception of recidivism risk (magnitude around +/- 0.15 on a 1-10 score scale). Note, however, that the set of Turk workers in the no-race and with-race conditions were different; this effect may diminish or exacerbate if the experiment is re-run with the same set of Turk workers.

\subsection{COMPAS + Humans: Characterizing Agreement \& Disagreement} 
We now determine the features that drive agreement or disagreement between COMPAS and Human scores.
To do so, we use two techniques -- clustering and decision trees. Specifically, we performed mean-shift clustering \cite{derpanis2005mean}, a robust-clustering method that avoids the need to specify an arbitrary number of clusters, to cluster defendants in each of Cases 1-8 from Table~\ref{tab:cases}. We also built a multiclass decision tree to classify individuals into each of the eight cases. Finally, we assessed the distribution of features across the found clusters and tree partitions. Figure \ref{fig:tree8} presents the decision tree. We elaborate on our findings below. A summary of the feature characteristics is in Table~\ref{tab:cases}.

\subsubsection{Easy calls: COMPAS and Humans agree, both correct.}
When we cluster defendants in this region of correct agreement, two clusters emerge that map to the two cases. The key separation between Cases 1 (COMPAS high, human high, both correct) and 2 (COMPAS low, Human low, both correct) is the number of priors, and to a lesser extent age. The average number of priors for defendants in Case $1$ is $7.9$, and $0.34$ for Case $2$. The average age for defendants in Case $1$ is $30.3$, and $40.56$ for Case $2$. Consequently, these cases correspond to what one might consider \textit{easy calls}, i.e., defendants for whom the number of priors and age alone provide sufficient information to predict recidivism accurately.

\subsubsection{Unknown unknowns: COMPAS and Humans agree, but both incorrect.} 
Now we turn our attention to the region of wrong agreement - defendants whose COMPAS and Human scores agree, yet fail to predict ground truth (Cases $7$ \& $8$). These defendants are very similar to defendants in other cases -- they are truly \emph{unknown unknowns}. Effectively, defendants in Cases 7 \& 8 are exactly defendants for whom the number of priors and age alone are not different enough to distinguish them from defendants in other cases, despite these defendants having fundamentally different ground truth labels. Because both COMPAS and Human scores are over reliant on the number of priors and age, both scores fail for defendants for whom these two features alone are not sufficient to predict recidivism.

\subsubsection{Characterizing the space of disagreement.} Our key finding for defendants for whom COMPAS and Human scores disagree (Cases 3-6) mirrors our findings for the unknown unknowns. These defendants had similar number of priors and age as defendants in other cases. In general, the four cases in the space of disagreement could not be cleanly separated from each other -- Cases 3 and 6 were similar; Cases 4 and 5 were similar -- and also overlapped with the space of agreement. For example, defendants with 
low COMPAS scores, high Human scores but did not recidivate (Case 4) tended to have 1.5 to 5.5 priors and are younger than 32.5 years old. However, these defendants significantly overlap with defendants in several other cases (Cases 1, 7, and 5 as seen in Table~\ref{tab:cases}). See Appendix B for details.

\subsection{COMPAS + Humans: Leveraging Disagreement to Build Hybrid Models}

Since defendants for whom COMPAS and Human scores disagree have the highest possibility of benefiting from hybrid models, we build two separate sets of hybrid models: (1) models on all defendants; (2) models on only the space of disagreement (32\% of defendants in this data). Table  \ref{tab:hybrid_disagree_all_reduced} reports the results of hybrid models trained and tested on only these defendants; results for the first set of hybrid models are in Appendix D. 
 
Hybrid methods tended to outperform single scores (or models trained on features and single scores) by a small margin. In Table \ref{tab:hybrid_disagree_all_reduced}, the best performing model (AUC 0.60) is a hybrid random forest predicting ground truth using features, COMPAS, and Human (no-race condition) scores. This was better than single risk scores (HNR 0.56, HWR 0.54, Compas 0.49), but comparable to a random forest model trained on the original features plus the HNR scores (but not with COMPAS), which obtained an AUC of 0.59.
Interestingly, despite the low AUC of COMPAS (0.49), combining it with HNR did not degrade the hybrid model's performance and in fact led to a small AUC improvement of 0.01. However, this improvement is within the margin of error.

Next, we examine these results by race. Table \ref{tab:hybrid_disagree_black} presents these
results for blacks, Table \ref{tab:hybrid_disagree_white} for whites, and Table \ref{tab:hybrid_disagree_other} for
other races. The trend is again similar, where hybrid models tended
to obtain slightly better results than their single-model counterparts, but improvements are typically within the margin of error. Hybrid models for blacks had the best accuracy and error rates; single models for other races (only 31 defendants) had the best accuracy and error rates. 

\begin{table}[ht]
\centering
\footnotesize
\caption{Test-set performance of hybrid models built on individuals whose COMPAS and Human risk scores disagree. Best results in cyan and bolded. See Table \ref{tab:hybrid_disagree_all} in the appendix for extended version of this table.}
\label{tab:hybrid_disagree_all_reduced}
\begin{tabular}{lll}
  \toprule
\textbf{Type} & \textbf{Model} & \textbf{AUC}  \\ 
  \toprule
 \multirow{3}{*}{Hybrid} & Best hybrid of C and HNR & \cellcolorandbold{Cyan}{0.60 $\pm$ 0.07}  \\ 
   & Best hybrid of C and HWR & 0.58 $\pm$ 0.08 \\ 
   & Best hybrid of C, HWR, HNR & 0.58 $\pm$ 0.07 \\ 
  \midrule
   \multirow{6}{*}{Single} & Predict GT from features and HNR & 0.59 $\pm$ 0.07\\ 
   & HNR (1-10 scale) & 0.56 $\pm$ 0.05 \\ 
   & Predict GT from features and HWR & 0.54 $\pm$ 0.06 \\ 
   & HWR (1-10 scale) & 0.54 $\pm$ 0.04\\ 
   & Predict GT from features and C & 0.51 $\pm$ 0.07 \\ 
   & C (1-10 scale) & 0.49 $\pm$ 0.06 \\ 
 \midrule
  None & Predict GT from features & 0.52 $\pm$ 0.07\\ 
  \midrule
 \multirow{3}{*}{Random} & Randomly pick between C and HNR & 0.55 $\pm$ 0.08  \\ 

   & Randomly pick between C and HWR & 0.54 $\pm$ 0.07\\ 
   & Randomly pick between C, HWR, HNR & 0.54 $\pm$ 0.06 \\ 
  \bottomrule
\end{tabular}
\end{table}

In general, as can be seen in Table \ref{tab:hybrid_disagree_all} to \ref{tab:hybrid_disagree_other}, the best hybrid models tended to leverage defendant and Human worker features, plus both risk scores, to either directly or indirectly predict ground truth recidivism. For the space of disagreement, the best hybrid models also tended to prefer HNR over HWR, particularly when
evaluating races other than whites. On the other hand, for the space of disagreement, hybrid
models based on (weighted) averages of the COMPAS and human
scores tend to underperform models that incorporated defendant and Human worker features. Notably, this is not the case for all defendants as the best performing hybrid models for all defendants were the optimally weighted average models (Table \ref{tab:hybrid_entire_all}). 

We have shown that for defendants for whom COMPAS and Human scores disagree,
hybrid models can be more beneficial than single risk scores (even when one of the scores is not as high performing as the other, as is the case for COMPAS compared to Humans for this set of individuals), but, in general, the improvements are marginal and, in many cases, within the margin of error.

\section{Discussion}
Our key finding is that \textbf{on this data set, COMPAS and Human decision making differed, but not in ways that could be leveraged to improve ground truth prediction.} From our analysis, the number of priors  is a key feature in both COMPAS and Human decision making. We saw that COMPAS and Humans tended to agree (and were right) on defendants with a very high or very low number of priors. We saw that the defendants that COMPAS and humans agreed on (but were wrong) were truly \textit{unknown unknowns} -- there was no discernable pattern in these cases. Unfortunately, they make up 19\% of the data, which bounds the maximal possible improvement from a hybrid model on this data. 

When we focused on the 32\% of defendants where COMPAS and Human decisions disagree, our hybrid models started to exhibit some improvement, though still within the margin of error.
The cases in this region were also the most uncertain, with single risk scores achieving between 0.49 and 0.56 AUC. We saw that for this region of uncertainty, single risk scores could be further improved by allowing them to see some amount of ground truth labels, alongside defendant features. We saw that number of priors, once again, and age were the two most important features to determine whether a defendant would fall in Case 3-6, although separation between these four cases was often not clear.  

Several reasons could explain why we were not getting better accuracy from the hybrid models: 1) Ground truth labels are noisy. 2) Turk workers are not experts. 3) Ground truth is inherently unpredictable or the features we have do not present enough information to predict ground truth accurately. 4) Small sample size.

\subsection{Noisy ground truth labels}
One limitation of our hybrid models is possible noise (or bias) in the ground truth labels in the ProPublica COMPAS data.  The ``primary" definition of recidivism from the US Sentencing Commission \shortcite{ussc2004measuring} is one of the following during the defendant's initial two years back in the community: (1) re-conviction for a new offence; (2) re-arrest with no conviction; (3) supervision revocation. Although this definition has traditionally been considered reliable, it is only a proxy for ground truth and does not cover defendants arrested but not convicted, or defendants not arrested despite committing crimes. Use of this definition is also susceptible to racial or socioeconomic bias, as people of color or those who live in poorer communities may experience higher levels of policing, resulting in a higher rates of re-arrests \cite{eckhouse2017big}. As we continue to develop machine learning models for recidivism, we need to reevaluate the ground truth labels we are collecting to ensure they are unbiased. 

\subsection{Criminal justice expertise}
It is important to note that the Human risk scores in our analyses were obtained from Mechanical Turk workers.
The ecological validity of using Turk workers may be low, as they have no criminal justice experience, and the decisions they are asked to make (whether a defendant recidivates or not) may have little relation to the types of decisions they make in their day-to-day lives. Gathering human data from judges, in actual legal settings, will help us further investigate the potential of hybrid models in fairness domains. We need to gather more quantitative and qualitative data on when judges and algorithmic systems agree and disagree, and what additional information the judge may be using to inform her decision. This could help hybrid models better discern when to choose human judgment over algorithmic prediction to achieve better performance overall.  

\subsection{Lacking evidence about the world}
We have two other hypotheses why our hybrid models only marginally improve over the accuracy of COMPAS or Human scores alone despite the presence of differences in COMPAS and Human reasoning. First, perhaps recidivism is an unpredictable event with a lot of inherent uncertainty, and as such, the accuracy of any model is limited. This is consistent with prior work that found similar AUCs for commercial recidivism prediction systems \cite{drake2014predicting}. Second, it could be that the seven features included in this data are not sufficient to properly evaluate recidivism risk. This second explanation is likely, since the Turk worker ratings are only based on those seven features (besides race). In a real world court setting, a judge has access to additional information that could be used to inform their reasoning. This ``private information'' may be helpful, however it remains to be seen if private information may also be detrimental to human reasoning, as seen in \citeauthor{Grove2000clinical} \shortcite{Grove2000clinical} and \citeauthor{kleinberg2017human} \shortcite{kleinberg2017human} where physicians and judges sometimes responded to private information in ways that caused them to deviate from optimal judgment. In general, tracking as many features as possible would enable more detailed study of the value of private information in complex decision making settings. Moreover, our analyses found that both COMPAS and Human both relied heavily on the number of priors and age. While these features may be considered ``objective'' (e.g. prior research showed a strong correlation between prior criminal record and recidivism \cite{ussc2016recidivism}), many defendants may appear similar when viewed through the lens of only two features.

\subsection{Small sample size}
\label{sec:smallsample}
In our hybrid models trained on only the 340 defendants for which COMPAS and Human scores disagreed, the improvements demonstrated were subsumed by large margin of errors. This was also the case for further subgroups of races (169 blacks, 114 whites, 31 other races). Repeating the Mechanical Turk experiment and hybrid models on a larger sample of the original ProPublica COMPAS data will provide more evidence as to whether human judgment can help machines in making recidivism predictions.

\section{Conclusion}
In complex settings, like a courtroom or hospital, it is unlikely that algorithmic systems will be making all decisions without input from human experts. Our approach focused efforts on cases where humans and machines disagree as a potential area to enhance decision making. Ultimately, we want to leverage the best of both worlds: humans that glean subtle, interpersonal insights from rich context, and machine algorithms that provide rigor and consistency. However, on this data set, human and machine decision making differed but not in ways that could be leveraged to improve ground truth prediction of recidivism. An important next step will be to further our investigation to include predictions made by judges in real-world settings. We hypothesize that the richness of the real-world may provide better context for enhanced hybrid Human + Machine models.

A key debate in recidivism predictions involves issues of bias and fairness, particularly for false positive and false negative judgments. Although our work uncovered a few aspects where race had an impact, it was not the primary focus of our work. We intend to look more closely at issues of bias and fairness in future work, especially as we gather more real-world data. Although both humans and algorithms can have inherent biases, if these biases are different, a hybrid model has the potential to help overcome them.

\clearpage

\bibliographystyle{aaai}
\bibliography{bibliography}

\clearpage
\setcounter{table}{0}
\renewcommand{\thetable}{A\arabic{table}}
\setcounter{figure}{0}
\renewcommand\thefigure{A\arabic{figure}}

\section*{Appendix A: Details of Hybrid Models and Metrics}

\subsubsection{Direct and indirect hybrid models.} 
\begin{figure}[hbt!]
\begin{center}
\includegraphics[width=0.35\textwidth]{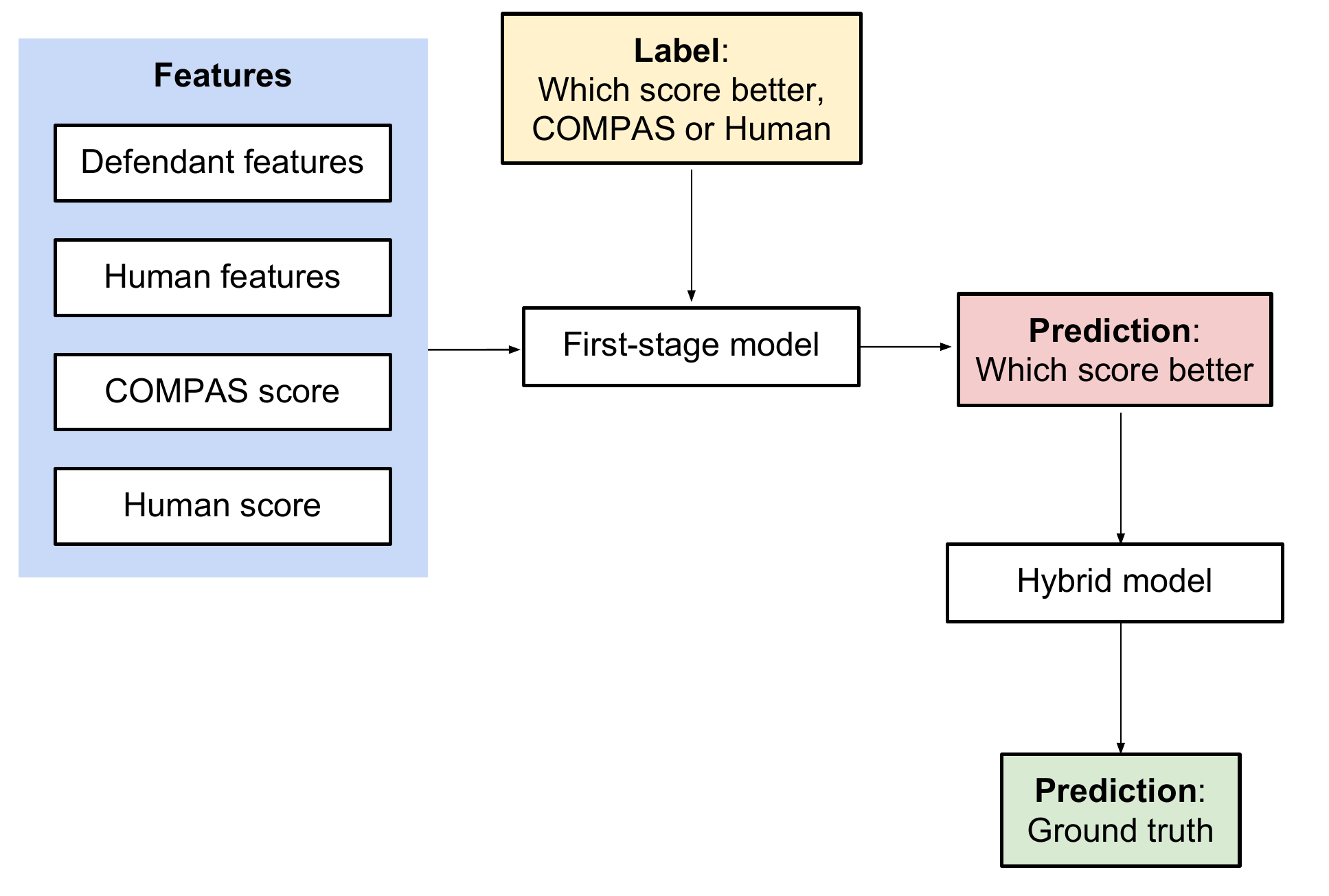} \\
\caption{Indirect hybrid model.}
\label{fig:indirecthybrid}
\end{center}
\end{figure}

\begin{figure}[hbt!]
\begin{center}
\includegraphics[width=0.35\textwidth]{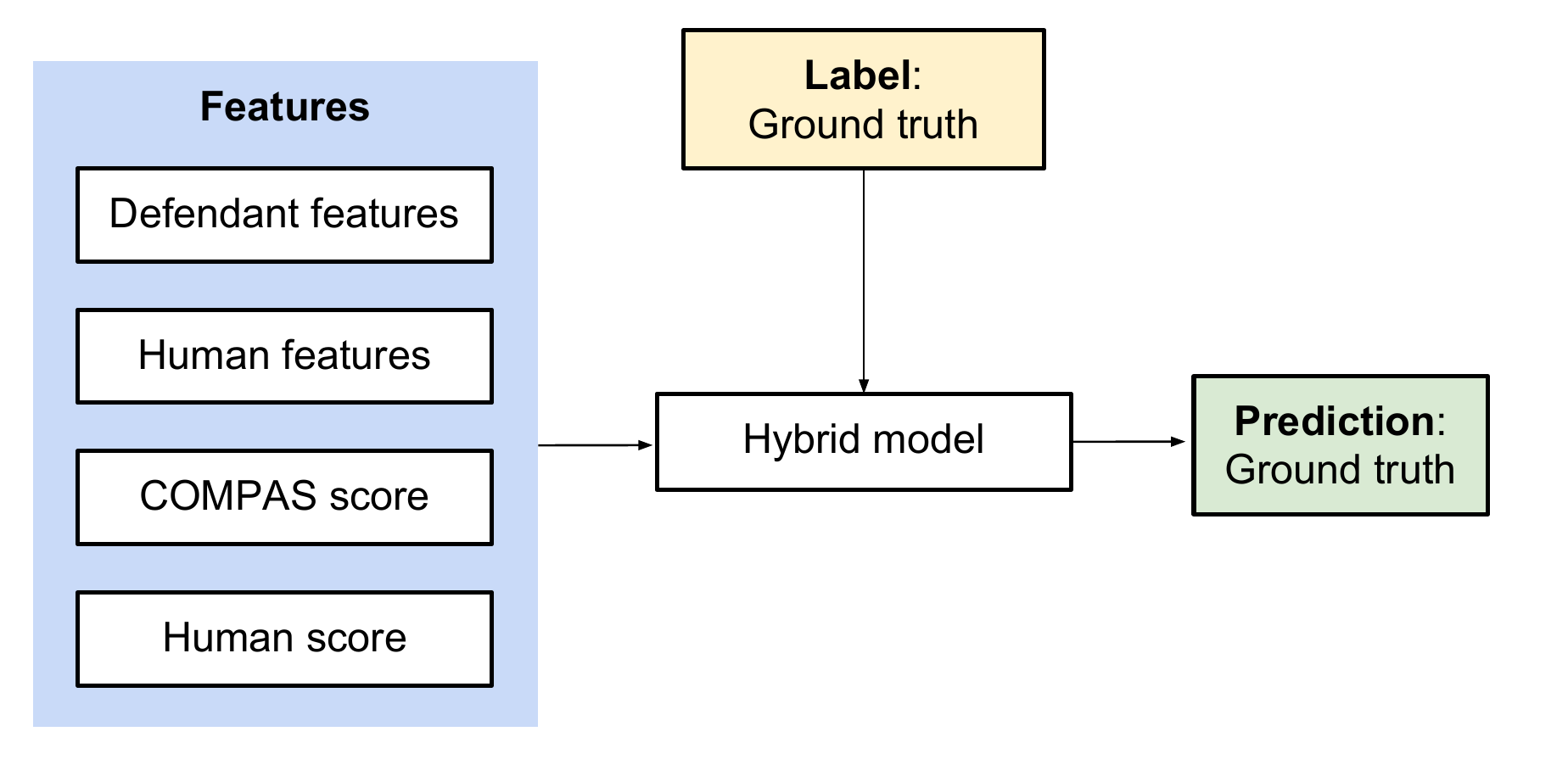} \\
\caption{Direct hybrid model.}
\label{fig:directhybrid}
\end{center}
\end{figure}

If we had access to an oracle that can be queried to obtain ground truth recidivism for any new observation, we can determine which of COMPAS or Human scores better predicts ground truth. However, test-time access to a ground truth oracle is not realistic. Hence, we relax this assumption of oracle access at test-time to only training-time, and train a binary classification model \textbf{only on observations where the two risk scores disagree} to predict which risk score to pick. In other words, this model predicts which score -- COMPAS or Human -- to use for Cases 3-6 in Table \ref{tab:cases} using features available at training time such as defendant features, Turk worker features, COMPAS score, and Human score. We call this an \textbf{indirect} hybrid model -- indirect because the hybrid model takes as input the prediction of which risk score is better, and outputs the desired ground truth recidivism prediction. Figure \ref{fig:indirecthybrid} shows this model. We also \textbf{directly} predict ground truth recidivism as a function of not just features but also the two risk scores. Figure \ref{fig:directhybrid} shows this model. 

\subsubsection{Accuracy and error metrics.} 
Besides AUC, we also report balanced accuracy (Bal Acc), i.e., the mean classification accuracy across classes. For error rates, we track false positives (FPR), false negatives (FNR), false discovery (FDR), and false omission (FOR). Equations for these error rates are below. Note that Kleinberg \cite{kleinberg2017inherent} and Choudechouva \cite{chouldechova2017fair} showed the impossibility of satisfying several of these metrics simultaneously.  

Given a binary label and a binary prediction, let FP denote the number of false positives, FN denote the number of false negatives, TP denote the number of true positives, and TN denote the number of true negatives.

\paragraph{Balanced accuracy} \begin{equation*} Bal Acc = \frac{1}{2}\left(\frac{TP}{TP+FN} + \frac{TN}{TN+FP}\right) \end{equation*}

\paragraph{False positive rate (FPR)} \begin{equation*} FPR = \frac{FP}{FP + TN} \end{equation*}

\paragraph{False negative rate (FPR)} \begin{equation*} FPR = \frac{FN}{FN + TP} \end{equation*}

\paragraph{False discovery rate (FDR)} \begin{equation*} FDR = \frac{FP}{FP + TP} \end{equation*}

\paragraph{False omission rate (FOR)} \begin{equation*} FOR = \frac{FN}{FN + TN} \end{equation*}

\subsection*{Appendix B: Detailed Characterization of the Space of Disagreement}

\paragraph{COMPAS score high, Human score low (Cases 3 \& 6).} The difference between Cases 3 and 6 is their ground truth label - defendants in Case 3 recidivated, whereas defendants in Case 6 did not. According to the decision tree's partitions, defendants in Cases $3$ and $6$ tend to have $<0.5$ priors. In fact, the key distinguishing feature between Cases 3 and 6 is the type of crime that the defendant was charged with. In addition, we found that some of the multiclass trees we built to predict classification into the eight cases did not always have terminal nodes with Case 6. Sometimes, Case 6 is combined with Case 3, indicating that the features do not have sufficient signal to adequately distinguish these two cases. 

\paragraph{COMPAS score low, Human score high (Cases 4 \& 5).} The difference between Cases 4 and 5 is also their ground truth label - defendants in Case 4 did not recidivate, whereas defendants in Case 5 did. 
Case 4 defendants tended to have 1.5 to 5.5 priors and be older than 32.5 years old. Case 5 was not always present as a terminal node in our trees, and are very similar to defendants in Case 4 and also Cases 1 and 7 (in the space of agreement). 

\clearpage 

\begin{sidewaysfigure*}[ht!]
\section*{Appendix B: Additional Figures}
\vspace{2cm}
\begin{center}
 \includegraphics[width=\textwidth]{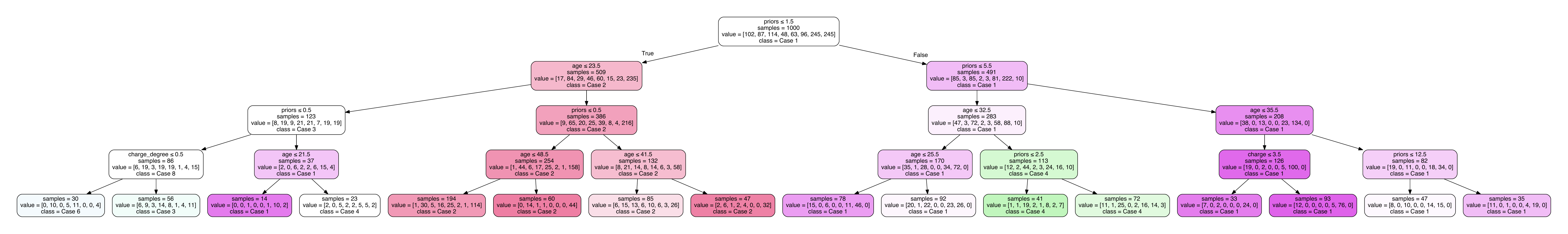}
\caption{Decision tree to explain the three-way interaction between COMPAS, Human scores, and ground truth. The label for the prediction task corresponds to the 8 different cases from Table \ref{tab:cases}. The five values in each node are (1) the partition of features defining that node (2) the number of samples in that node (3) the number of samples in each of the 8 cases (4) the predicted case (of the 8 cases). Best viewed after downloading and zooming in a PDF reader.}
\label{fig:tree8}
\end{center}
\vspace{-2cm}
\end{sidewaysfigure*}

\begin{figure*}[hbt!]
\begin{center}
\begin{tabular}{c}
\includegraphics[width=0.33\textwidth]{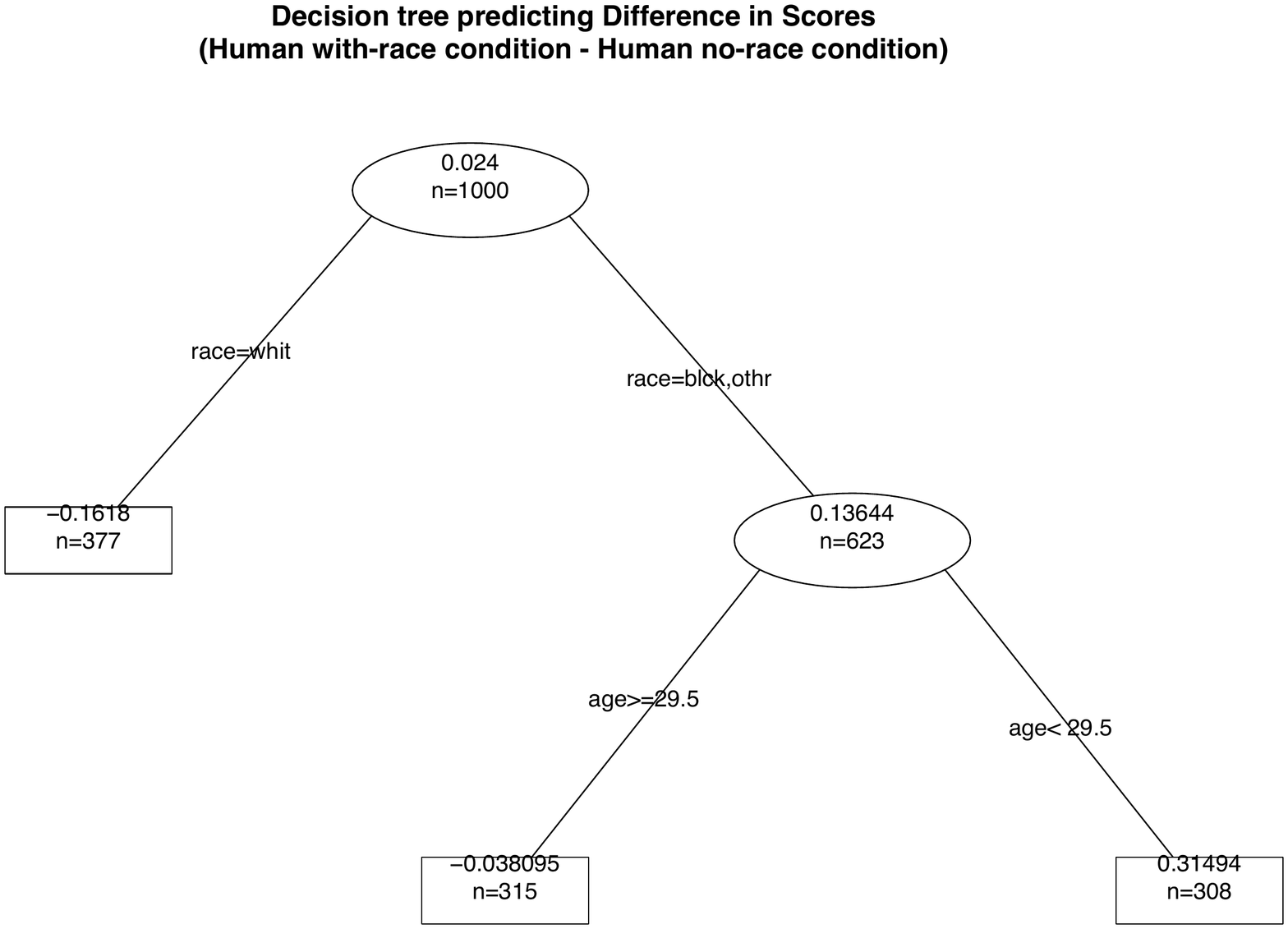}
\includegraphics[width=0.33\textwidth]{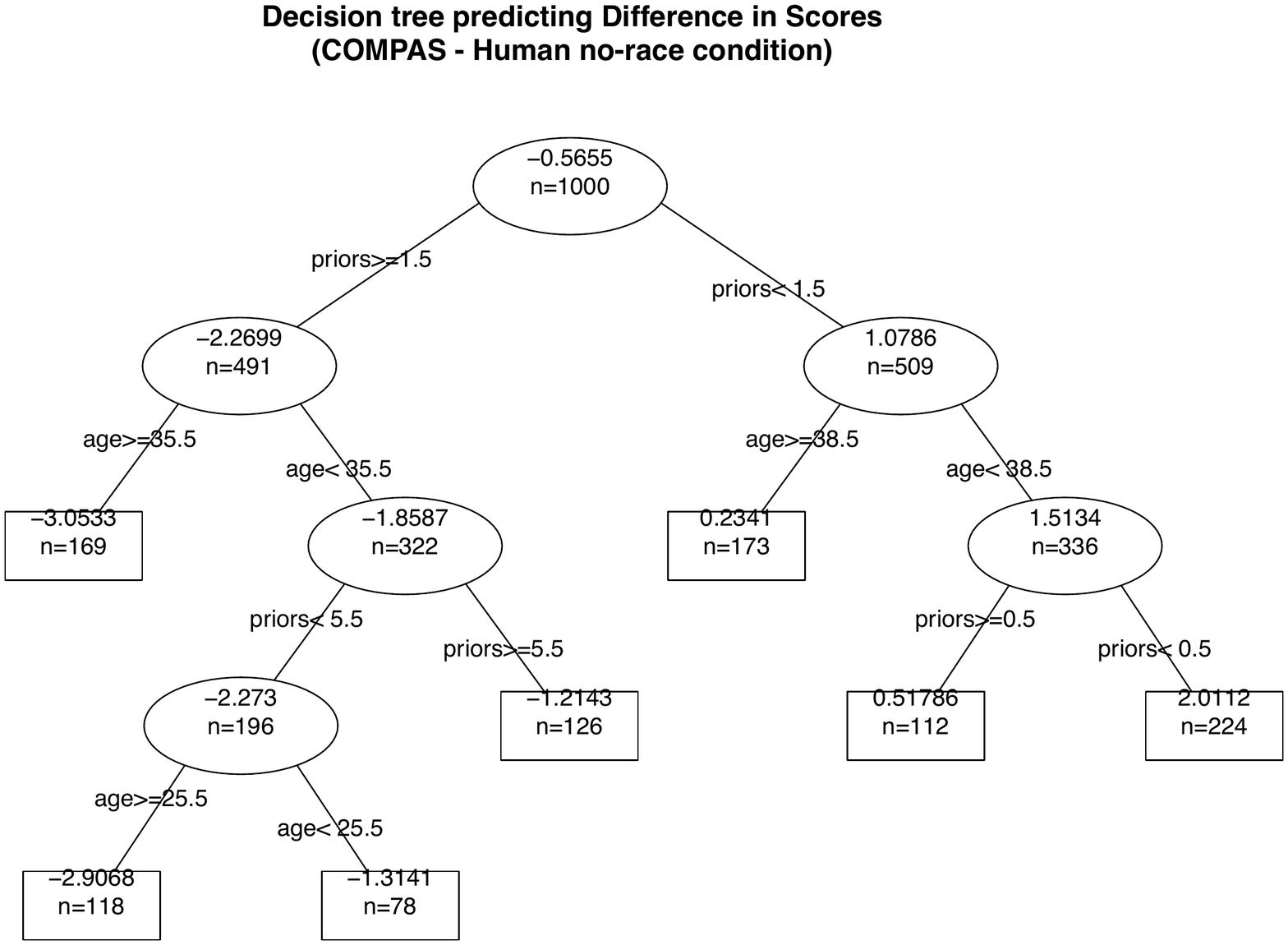}
\includegraphics[width=0.33\textwidth]{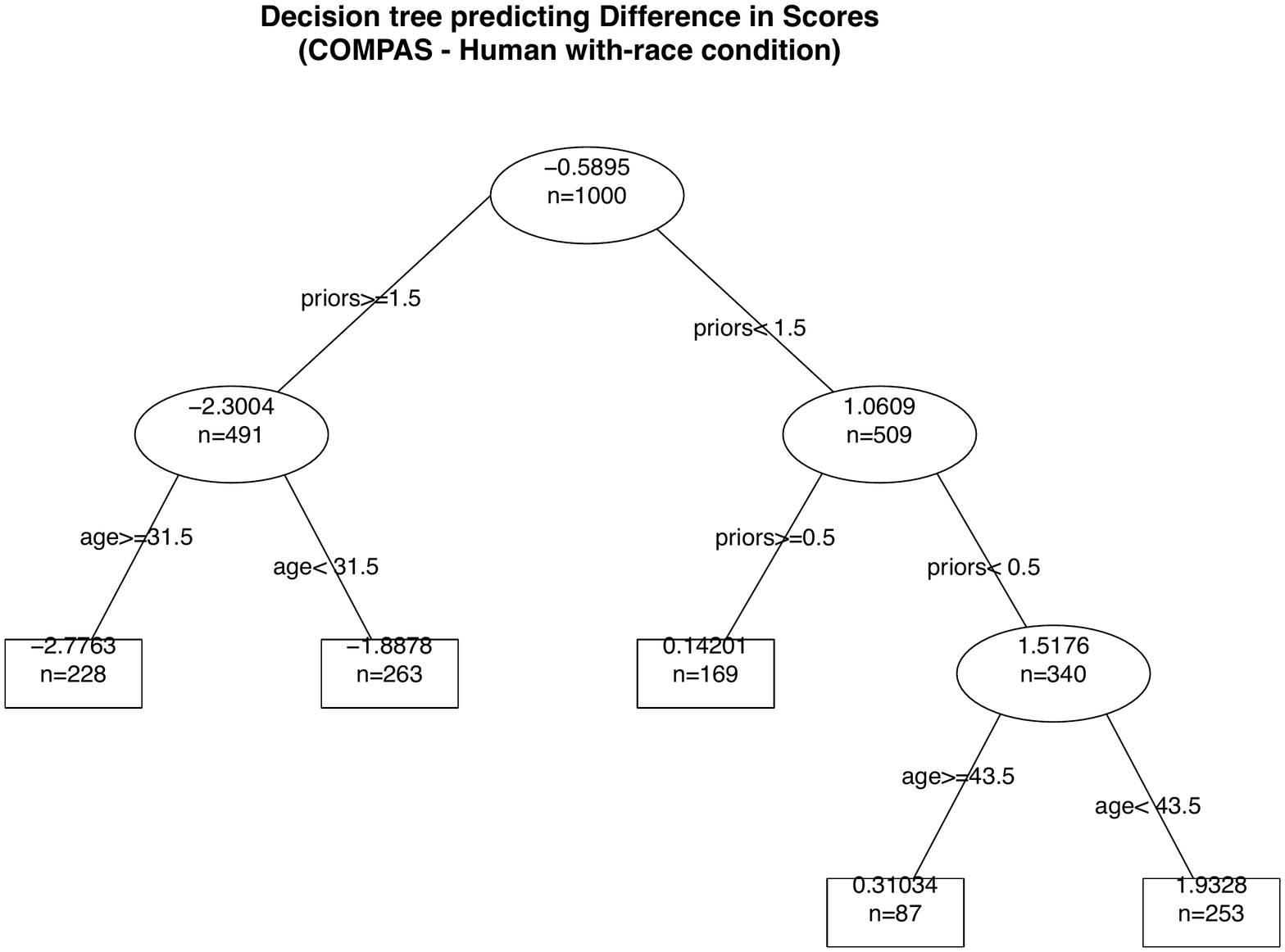}
\end{tabular}
\caption{Decision tree predicting the difference between scores. Left: difference in scores given by Turk workers when and when not told of the defendant's race (HWR - HNR). Center: difference in scores given by COMPAS and Turk workers not told of the defendant's race (C - HNR). Right:difference in scores given by COMPAS and Turk workers told of the defendant's race (C - HWR).}
\label{fig:tree3}
\end{center}
\end{figure*}

\begin{figure*}[htb]
\begin{center}
\begin{tabular}{ccc}
\includegraphics[width=0.33\textwidth,height=1.4in]{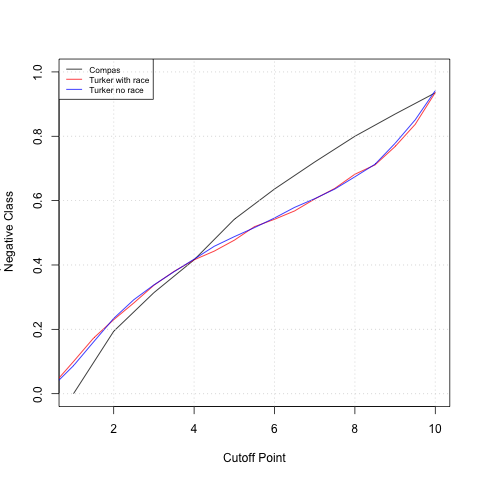}
\includegraphics[width=0.33\textwidth,height=1.4in]{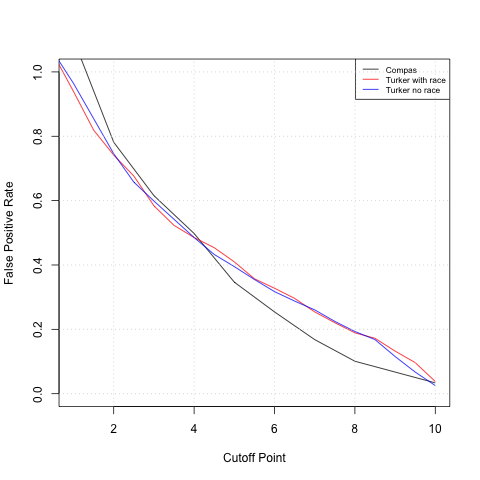} 
\includegraphics[width=0.33\textwidth,height=1.4in]{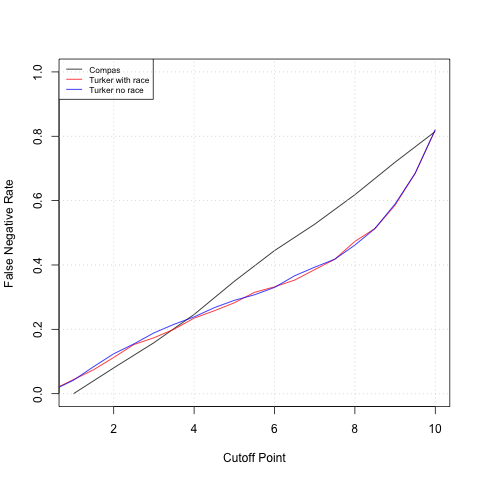} 
\end{tabular}
\caption{Accuracies (left), false positive rates (center), and false negative rates (right) for COMPAS and Human scores at different cutoff points for binarizing the scores.}
\label{fig:calibration}
\end{center}
\end{figure*}

\clearpage

\begin{table*}[ht]
\section*{Appendix C: Extended result tables for hybrid models for defendants whose \\ COMPAS and Human scores disagree}
\centering
\footnotesize
\caption{Test-set performance of hybrid models built on individuals whose COMPAS and Human risk scores disagree, compared to just using a single risk score and other baselines. The numbers presented are means and standard deviations calculated over 10 train-test splits.  Best results in cyan and bolded. A reduced version of this table can be seen in Table \ref{tab:hybrid_disagree_all_reduced}. Rows marked with $^*$ are the rows labeled as \emph{best} in Table \ref{tab:hybrid_disagree_all_reduced}.}
\label{tab:hybrid_disagree_all}
\vspace{0.25cm}
\begin{tabular}{llllllll}
  \toprule
\textbf{Type} & \textbf{Model} & \textbf{AUC} & \textbf{Bal Acc} & \textbf{FPR} & \textbf{FNR} & \textbf{FDR} & \textbf{FOR} \\ 
  \toprule
Hybrid & Direct C HNR$^*$ & \cellcolorandbold{Cyan}{0.60 $\pm$ 0.07} & \cellcolorandbold{Cyan}{0.56 $\pm$ 0.07} & 0.44 $\pm$ 0.13 & 0.45 $\pm$ 0.10& 0.50 $\pm$ 0.10& \cellcolorandbold{Cyan}{0.39 $\pm$ 0.08} \\ 
  Hybrid & Composed indirect C HWR$^*$ & 0.58 $\pm$ 0.08 & \cellcolorandbold{Cyan}{0.56 $\pm$ 0.08} & 0.37 $\pm$ 0.10& 0.50 $\pm$ 0.10& \cellcolorandbold{Cyan}{0.47 $\pm$ 0.15} & 0.40 $\pm$ 0.10\\ 
  Hybrid & Direct C HWR HNR$^*$ & 0.58 $\pm$ 0.07 & 0.55 $\pm$ 0.08 & 0.47 $\pm$ 0.14 & 0.43 $\pm$ 0.09 & 0.50 $\pm$ 0.09 & 0.40 $\pm$ 0.10\\ 
  Hybrid & Indirect C HWR$^*$ & 0.58 $\pm$ 0.08 & \cellcolorandbold{Cyan}{0.56 $\pm$ 0.08} & 0.37 $\pm$ 0.10& 0.50 $\pm$ 0.10& \cellcolorandbold{Cyan}{0.47 $\pm$ 0.15} & 0.40 $\pm$ 0.10\\ 
  Hybrid & Composed indirect C HNR & 0.56 $\pm$ 0.09 & 0.54 $\pm$ 0.06 & 0.45 $\pm$ 0.07 & 0.47 $\pm$ 0.09 & 0.52 $\pm$ 0.07 & 0.40 $\pm$ 0.08 \\ 
  Hybrid & Indirect C HNR & 0.56 $\pm$ 0.09 & 0.54 $\pm$ 0.06 & 0.45 $\pm$ 0.07 & 0.47 $\pm$ 0.09 & 0.52 $\pm$ 0.07 & 0.40 $\pm$ 0.08 \\ 
  Hybrid & Direct C HWR & 0.53 $\pm$ 0.06 & 0.52 $\pm$ 0.04 & 0.37 $\pm$ 0.09 & 0.58 $\pm$ 0.09 & 0.52 $\pm$ 0.14 & 0.44 $\pm$ 0.08 \\ 
  Hybrid & Weighted average of C HNR & 0.51 $\pm$ 0.05 & 0.50 $\pm$ 0.04 & 0.38 $\pm$ 0.25 & 0.63 $\pm$ 0.3 & 0.56 $\pm$ 0.22 & 0.43 $\pm$ 0.07 \\ 
  Hybrid & Weighted average of C HWR HNR & 0.50 $\pm$ 0.04 & 0.50 $\pm$ 0.05 & \cellcolorandbold{Cyan}{0.23 $\pm$ 0.07} & 0.77 $\pm$ 0.13 & 0.58 $\pm$ 0.09 & 0.45 $\pm$ 0.11 \\ 
  Hybrid & Weighted average of C HWR & 0.47 $\pm$ 0.04 & 0.49 $\pm$ 0.03 & 0.39 $\pm$ 0.26 & 0.63 $\pm$ 0.26 & 0.56 $\pm$ 0.12 & 0.46 $\pm$ 0.11 \\ 
  \midrule
  Single & Predict GT from features and HNR & 0.59 $\pm$ 0.07 & 0.55 $\pm$ 0.06 & 0.44 $\pm$ 0.09 & 0.46 $\pm$ 0.10& 0.51 $\pm$ 0.09 & \cellcolorandbold{Cyan}{0.39 $\pm$ 0.07} \\ 
  Single & HNR (1-10 scale) & 0.56 $\pm$ 0.05 & 0.52 $\pm$ 0.02 & 0.55 $\pm$ 0.08 & 0.40 $\pm$ 0.08 & 0.54 $\pm$ 0.04 & 0.41 $\pm$ 0.07 \\ 
  Single & Predict GT from features and HWR & 0.54 $\pm$ 0.06 & 0.54 $\pm$ 0.05 & 0.35 $\pm$ 0.10& 0.57 $\pm$ 0.08 & 0.49 $\pm$ 0.14 & 0.42 $\pm$ 0.09 \\ 
  Single & HWR (1-10 scale) & 0.54 $\pm$ 0.04 & 0.52 $\pm$ 0.03 & 0.54 $\pm$ 0.05 & 0.41 $\pm$ 0.04 & 0.53 $\pm$ 0.09 & 0.43 $\pm$ 0.10\\ 
  Single & Predict GT from features and C & 0.51 $\pm$ 0.07 & 0.52 $\pm$ 0.05 & 0.41 $\pm$ 0.07 & 0.55 $\pm$ 0.08 & 0.52 $\pm$ 0.11 & 0.44 $\pm$ 0.10\\ 
  Single & C (1-10 scale) & 0.49 $\pm$ 0.06 & 0.48 $\pm$ 0.01 & 0.40 $\pm$ 0.07 & 0.65 $\pm$ 0.08 & 0.59 $\pm$ 0.06 & 0.46 $\pm$ 0.04 \\ 
  Single & C (binarized $>$=5) & - & 0.48 $\pm$ 0.01 & 0.40 $\pm$ 0.07 & 0.65 $\pm$ 0.08 & 0.59 $\pm$ 0.06 & 0.46 $\pm$ 0.04 \\ 
  Single & HNR (binarized $>$=5) & - & 0.52 $\pm$ 0.01 & 0.60 $\pm$ 0.07 & \cellcolorandbold{Cyan}{0.35 $\pm$ 0.08} & 0.54 $\pm$ 0.04 & 0.41 $\pm$ 0.06 \\ 
  Single & HWR (binarized $>$=5) & - & 0.51 $\pm$ 0.03 & 0.63 $\pm$ 0.05 & 0.36 $\pm$ 0.05 & 0.54 $\pm$ 0.07 & 0.44 $\pm$ 0.12 \\ 
 \midrule
  None & Predict GT from features & 0.52 $\pm$ 0.07 & 0.51 $\pm$ 0.06 & 0.37 $\pm$ 0.07 & 0.61 $\pm$ 0.09 & 0.54 $\pm$ 0.13 & 0.45 $\pm$ 0.09 \\ 
  \midrule
  Random & Randomly pick between C HNR & 0.55 $\pm$ 0.08 & 0.52 $\pm$ 0.05 & 0.46 $\pm$ 0.06 & 0.50 $\pm$ 0.08 & 0.54 $\pm$ 0.07 & 0.42 $\pm$ 0.07 \\ 
  Random & Randomly pick between C HWR & 0.54 $\pm$ 0.07 & 0.52 $\pm$ 0.06 & 0.46 $\pm$ 0.06 & 0.49 $\pm$ 0.11 & 0.53 $\pm$ 0.14 & 0.43 $\pm$ 0.08 \\ 
  Random & Randomly pick between C HWR HNR & 0.54 $\pm$ 0.06 & 0.52 $\pm$ 0.06 & 0.50 $\pm$ 0.07 & 0.46 $\pm$ 0.08 & 0.53 $\pm$ 0.10& 0.43 $\pm$ 0.11 \\ 
  \bottomrule
\end{tabular}
\end{table*}

\begin{table*}[ht]
\centering
\footnotesize
\caption{Subgroup (African-Americans) performance of models presented in Table \ref{tab:hybrid_disagree_all}.}
\label{tab:hybrid_disagree_black}
\begin{tabular}{llllllll}
\toprule
\textbf{Type} & \textbf{Model} & \textbf{AUC} & \textbf{Bal Acc} & \textbf{FPR} & \textbf{FNR} & \textbf{FDR} & \textbf{FOR} \\ 
\toprule
Hybrid & Direct C HNR & \cellcolorandbold{Cyan}{0.65 $\pm$ 0.06} & 0.58 $\pm$ 0.07 & 0.48 $\pm$ 0.15 & 0.35 $\pm$ 0.13 & 0.46 $\pm$ 0.11 & 0.37 $\pm$ 0.10\\ 
  Hybrid & Direct C HWR HNR & 0.63 $\pm$ 0.07 & \cellcolorandbold{Cyan}{0.59 $\pm$ 0.07} & 0.50 $\pm$ 0.15 & \cellcolorandbold{Cyan}{0.32 $\pm$ 0.11} & 0.45 $\pm$ 0.09 & \cellcolorandbold{Cyan}{0.36 $\pm$ 0.13} \\ 
  Hybrid & Composed indirect C HWR & 0.57 $\pm$ 0.12 & 0.58 $\pm$ 0.10& 0.41 $\pm$ 0.15 & 0.43 $\pm$ 0.14 & 0.43 $\pm$ 0.17 & 0.41 $\pm$ 0.12 \\ 
  Hybrid & Indirect C HWR & 0.57 $\pm$ 0.12 & 0.58 $\pm$ 0.10& 0.41 $\pm$ 0.15 & 0.43 $\pm$ 0.14 & 0.43 $\pm$ 0.17 & 0.41 $\pm$ 0.12 \\ 
  Hybrid & Weighted average of C HNR & 0.55 $\pm$ 0.06 & 0.51 $\pm$ 0.07 & 0.33 $\pm$ 0.19 & 0.64 $\pm$ 0.31 & 0.65 $\pm$ 0.21 & 0.42 $\pm$ 0.12 \\ 
  Hybrid & Weighted average of C HWR HNR & 0.55 $\pm$ 0.08 & 0.55 $\pm$ 0.08 & \cellcolorandbold{Cyan}{0.12 $\pm$ 0.09} & 0.78 $\pm$ 0.15 & \cellcolorandbold{Cyan}{0.35 $\pm$ 0.21} & 0.45 $\pm$ 0.15 \\ 
  Hybrid & Composed indirect C HNR & 0.53 $\pm$ 0.09 & 0.52 $\pm$ 0.07 & 0.54 $\pm$ 0.10& 0.41 $\pm$ 0.07 & 0.52 $\pm$ 0.07 & 0.44 $\pm$ 0.13 \\ 
  Hybrid & Indirect C HNR & 0.53 $\pm$ 0.09 & 0.52 $\pm$ 0.07 & 0.54 $\pm$ 0.10& 0.41 $\pm$ 0.07 & 0.52 $\pm$ 0.07 & 0.44 $\pm$ 0.13 \\ 
  Hybrid & Direct C HWR & 0.51 $\pm$ 0.07 & 0.51 $\pm$ 0.07 & 0.40 $\pm$ 0.14 & 0.57 $\pm$ 0.12 & 0.50 $\pm$ 0.17 & 0.48 $\pm$ 0.10\\ 
  Hybrid & Weighted average of C HWR & 0.48 $\pm$ 0.09 & 0.49 $\pm$ 0.07 & 0.37 $\pm$ 0.27 & 0.65 $\pm$ 0.26 & 0.50 $\pm$ 0.16 & 0.51 $\pm$ 0.15 \\ 
  \midrule
  Single & Predict GT from features and HNR & 0.64 $\pm$ 0.06 & 0.56 $\pm$ 0.06 & 0.48 $\pm$ 0.13 & 0.39 $\pm$ 0.13 & 0.48 $\pm$ 0.10& 0.39 $\pm$ 0.09 \\ 
  Single & HNR (1-10 scale) & 0.55 $\pm$ 0.07 & 0.56 $\pm$ 0.05 & 0.46 $\pm$ 0.09 & 0.42 $\pm$ 0.15 & 0.49 $\pm$ 0.08 & 0.39 $\pm$ 0.09 \\ 
  Single & HWR (1-10 scale) & 0.53 $\pm$ 0.08 & 0.53 $\pm$ 0.06 & 0.47 $\pm$ 0.08 & 0.47 $\pm$ 0.11 & 0.49 $\pm$ 0.13 & 0.46 $\pm$ 0.13 \\ 
  Single & Predict GT from features and HWR & 0.52 $\pm$ 0.08 & 0.53 $\pm$ 0.08 & 0.36 $\pm$ 0.15 & 0.57 $\pm$ 0.13 & 0.46 $\pm$ 0.17 & 0.46 $\pm$ 0.12 \\ 
  Single & Predict GT from features and C & 0.49 $\pm$ 0.11 & 0.50 $\pm$ 0.06 & 0.48 $\pm$ 0.12 & 0.52 $\pm$ 0.13 & 0.52 $\pm$ 0.13 & 0.49 $\pm$ 0.12 \\ 
  Single & C (1-10 scale) & 0.46 $\pm$ 0.06 & 0.44 $\pm$ 0.05 & 0.51 $\pm$ 0.10& 0.60 $\pm$ 0.16 & 0.61 $\pm$ 0.09 & 0.51 $\pm$ 0.08 \\ 
  Single & C (binarized $>$=5) & - & 0.44 $\pm$ 0.05 & 0.51 $\pm$ 0.10& 0.60 $\pm$ 0.16 & 0.61 $\pm$ 0.09 & 0.51 $\pm$ 0.08 \\ 
  Single & HNR (binarized $>$=5) & - & 0.56 $\pm$ 0.05 & 0.49 $\pm$ 0.10& 0.40 $\pm$ 0.16 & 0.49 $\pm$ 0.08 & 0.39 $\pm$ 0.09 \\ 
  Single & HWR (binarized $>$=5) & - & 0.51 $\pm$ 0.06 & 0.57 $\pm$ 0.11 & 0.42 $\pm$ 0.12 & 0.51 $\pm$ 0.11 & 0.48 $\pm$ 0.16 \\
  \midrule
  None & Predict GT from features & 0.49 $\pm$ 0.10& 0.48 $\pm$ 0.10& 0.42 $\pm$ 0.13 & 0.62 $\pm$ 0.12 & 0.54 $\pm$ 0.19 & 0.50 $\pm$ 0.11 \\ 
  \midrule
  Random & Randomly pick between C HWR & 0.59 $\pm$ 0.06 & 0.57 $\pm$ 0.07 & 0.47 $\pm$ 0.08 & 0.40 $\pm$ 0.13 & 0.46 $\pm$ 0.14 & 0.41 $\pm$ 0.11 \\ 
  Random & Randomly pick between C HWR HNR & 0.54 $\pm$ 0.07 & 0.53 $\pm$ 0.06 & 0.48 $\pm$ 0.09 & 0.47 $\pm$ 0.11 & 0.49 $\pm$ 0.11 & 0.46 $\pm$ 0.13 \\ 
  Random & Randomly pick between C HNR & 0.53 $\pm$ 0.13 & 0.50 $\pm$ 0.10& 0.53 $\pm$ 0.11 & 0.47 $\pm$ 0.11 & 0.54 $\pm$ 0.07 & 0.46 $\pm$ 0.15 \\ 
   \bottomrule
\end{tabular}
\end{table*}

\begin{table*}[ht]
\centering
\footnotesize
\caption{Subgroup (whites) performance of models presented in Table \ref{tab:hybrid_disagree_all}.}
\label{tab:hybrid_disagree_white}
\begin{tabular}{llllllll}
  \hline
\textbf{Type} & \textbf{Model} & \textbf{AUC} & \textbf{Bal Acc} & \textbf{FPR} & \textbf{FNR} & \textbf{FDR} & \textbf{FOR} \\
  \hline
Hybrid & Composed indirect C HWR & 0.59 $\pm$ 0.09 & 0.55 $\pm$ 0.08 & \cellcolorandbold{Cyan}{0.30 $\pm$ 0.14} & 0.61 $\pm$ 0.11 & 0.49 $\pm$ 0.21 & 0.39 $\pm$ 0.10\\ 
  Hybrid & Indirect C HWR & 0.59 $\pm$ 0.09 & 0.55 $\pm$ 0.08 & \cellcolorandbold{Cyan}{0.30 $\pm$ 0.14} & 0.61 $\pm$ 0.11 & 0.49 $\pm$ 0.21 & 0.39 $\pm$ 0.10\\ 
  Hybrid & Composed indirect C HNR & 0.56 $\pm$ 0.2 & 0.52 $\pm$ 0.17 & 0.41 $\pm$ 0.16 & 0.56 $\pm$ 0.2 & 0.55 $\pm$ 0.19 & 0.42 $\pm$ 0.18 \\ 
  Hybrid & Direct C HWR & 0.56 $\pm$ 0.08 & \cellcolorandbold{Cyan}{0.58 $\pm$ 0.09} & 0.31 $\pm$ 0.18 & 0.53 $\pm$ 0.17 & \cellcolorandbold{Cyan}{0.45 $\pm$ 0.24} & \cellcolorandbold{Cyan}{0.36 $\pm$ 0.12} \\ 
  Hybrid & Indirect C HNR & 0.56 $\pm$ 0.2 & 0.52 $\pm$ 0.17 & 0.41 $\pm$ 0.16 & 0.56 $\pm$ 0.2 & 0.55 $\pm$ 0.19 & 0.42 $\pm$ 0.18 \\ 
  Hybrid & Direct C HNR & 0.53 $\pm$ 0.17 & 0.52 $\pm$ 0.13 & 0.39 $\pm$ 0.11 & 0.57 $\pm$ 0.21 & 0.56 $\pm$ 0.17 & 0.42 $\pm$ 0.15 \\ 
  Hybrid & Direct C HWR HNR & 0.52 $\pm$ 0.19 & 0.50 $\pm$ 0.13 & 0.43 $\pm$ 0.14 & 0.56 $\pm$ 0.2 & 0.57 $\pm$ 0.17 & 0.44 $\pm$ 0.14 \\ 
  Hybrid & Weighted average of C HWR & 0.48 $\pm$ 0.11 & 0.48 $\pm$ 0.06 & 0.42 $\pm$ 0.28 & 0.61 $\pm$ 0.25 & 0.59 $\pm$ 0.19 & 0.45 $\pm$ 0.14 \\ 
  Hybrid & Weighted average of C HWR HNR & 0.46 $\pm$ 0.08 & 0.44 $\pm$ 0.05 & 0.36 $\pm$ 0.10& 0.76 $\pm$ 0.19 & 0.72 $\pm$ 0.13 & 0.46 $\pm$ 0.11 \\ 
  Hybrid & Weighted average of C HNR & 0.44 $\pm$ 0.08 & 0.46 $\pm$ 0.06 & 0.47 $\pm$ 0.33 & 0.60 $\pm$ 0.3 & 0.55 $\pm$ 0.21 & 0.51 $\pm$ 0.21 \\ 
  \midrule
  Single & Predict GT from features and HWR & 0.59 $\pm$ 0.09 & \cellcolorandbold{Cyan}{0.58 $\pm$ 0.08} & 0.32 $\pm$ 0.15 & 0.52 $\pm$ 0.12 & 0.46 $\pm$ 0.21 & \cellcolorandbold{Cyan}{0.36 $\pm$ 0.1} \\ 
  Single & Predict GT from features and C & 0.56 $\pm$ 0.14 & 0.55 $\pm$ 0.12 & 0.34 $\pm$ 0.10& 0.56 $\pm$ 0.23 & 0.53 $\pm$ 0.16 & 0.39 $\pm$ 0.14 \\ 
  Single & HWR (1-10 scale) & 0.56 $\pm$ 0.12 & 0.53 $\pm$ 0.11 & 0.60 $\pm$ 0.18 & 0.34 $\pm$ 0.18 & 0.55 $\pm$ 0.12 & 0.39 $\pm$ 0.22 \\ 
  Single & Predict GT from features and HNR & 0.53 $\pm$ 0.19 & 0.53 $\pm$ 0.15 & 0.41 $\pm$ 0.12 & 0.54 $\pm$ 0.22 & 0.55 $\pm$ 0.17 & 0.41 $\pm$ 0.17 \\ 
  Single & HNR (1-10 scale) & 0.53 $\pm$ 0.15 & 0.46 $\pm$ 0.10& 0.67 $\pm$ 0.15 & 0.42 $\pm$ 0.16 & 0.59 $\pm$ 0.09 & 0.51 $\pm$ 0.25 \\ 
  Single & C (1-10 scale) & 0.52 $\pm$ 0.11 & 0.48 $\pm$ 0.09 & 0.35 $\pm$ 0.15 & 0.69 $\pm$ 0.17 & 0.60 $\pm$ 0.21 & 0.44 $\pm$ 0.10\\ 
  Single & C (binarized $>$=5) & - & 0.48 $\pm$ 0.09 & 0.35 $\pm$ 0.15 & 0.69 $\pm$ 0.17 & 0.60 $\pm$ 0.21 & 0.44 $\pm$ 0.10\\ 
  Single & HNR (binarized $>$=5) & - & 0.47 $\pm$ 0.07 & 0.72 $\pm$ 0.13 & 0.34 $\pm$ 0.14 & 0.58 $\pm$ 0.08 & 0.50 $\pm$ 0.24 \\ 
  Single & HWR (binarized $>$=5) & - & 0.52 $\pm$ 0.09 & 0.65 $\pm$ 0.15 & \cellcolorandbold{Cyan}{0.31 $\pm$ 0.17} & 0.56 $\pm$ 0.10& 0.40 $\pm$ 0.21 \\ 
  \midrule
  None & Predict GT from features & 0.55 $\pm$ 0.14 & 0.55 $\pm$ 0.12 & 0.37 $\pm$ 0.11 & 0.53 $\pm$ 0.2 & 0.51 $\pm$ 0.16 & 0.39 $\pm$ 0.14 \\ 
  \midrule
  Random & Randomly pick between C HNR & \cellcolorandbold{Cyan}{0.61 $\pm$ 0.14} & 0.57 $\pm$ 0.13 & 0.37 $\pm$ 0.11 & 0.49 $\pm$ 0.18 & 0.49 $\pm$ 0.14 & 0.38 $\pm$ 0.17 \\ 
  Random & Randomly pick between C HWR HNR & 0.54 $\pm$ 0.12 & 0.50 $\pm$ 0.14 & 0.54 $\pm$ 0.13 & 0.45 $\pm$ 0.18 & 0.57 $\pm$ 0.14 & 0.42 $\pm$ 0.17 \\ 
  Random & Randomly pick between C HWR & 0.49 $\pm$ 0.08 & 0.47 $\pm$ 0.08 & 0.47 $\pm$ 0.12 & 0.58 $\pm$ 0.13 & 0.61 $\pm$ 0.17 & 0.45 $\pm$ 0.09 \\ 
  \bottomrule
\end{tabular}
\end{table*}

\begin{table*}[ht]
\centering
\footnotesize
\caption{Subgroup (other races) performance of models presented in Table \ref{tab:hybrid_disagree_all}.}
\label{tab:hybrid_disagree_other}
\begin{tabular}{llllllll}
  \toprule
\textbf{Type} & \textbf{Model} & \textbf{AUC} & \textbf{Bal Acc} & \textbf{FPR} & \textbf{FNR} & \textbf{FDR} & \textbf{FOR} \\
  \toprule
Hybrid & Weighted average of C HNR & 0.64 $\pm$ 0.33 & 0.58 $\pm$ 0.23 & 0.47 $\pm$ 0.33 & 0.38 $\pm$ 0.44 & 0.63 $\pm$ 0.34 & 0.22 $\pm$ 0.25 \\ 
  Hybrid & Composed indirect C HNR & 0.62 $\pm$ 0.32 & 0.59 $\pm$ 0.24 & 0.29 $\pm$ 0.26 & 0.53 $\pm$ 0.39 & 0.52 $\pm$ 0.38 & 0.29 $\pm$ 0.21 \\ 
  Hybrid & Indirect C HNR & 0.62 $\pm$ 0.32 & 0.59 $\pm$ 0.24 & 0.29 $\pm$ 0.26 & 0.53 $\pm$ 0.39 & 0.52 $\pm$ 0.38 & 0.29 $\pm$ 0.21 \\ 
  Hybrid & Weighted average of C HWR HNR & 0.50 $\pm$ 0.25 & 0.57 $\pm$ 0.21 & 0.58 $\pm$ 0.37 & 0.29 $\pm$ 0.3 & 0.52 $\pm$ 0.27 & 0.33 $\pm$ 0.37 \\ 
  Hybrid & Weighted average of C HWR & 0.49 $\pm$ 0.24 & 0.52 $\pm$ 0.14 & 0.64 $\pm$ 0.26 & 0.32 $\pm$ 0.39 & 0.59 $\pm$ 0.23 & 0.29 $\pm$ 0.37 \\ 
  Hybrid & Direct C HWR HNR & 0.46 $\pm$ 0.22 & 0.44 $\pm$ 0.22 & 0.47 $\pm$ 0.31 & 0.66 $\pm$ 0.35 & 0.74 $\pm$ 0.33 & 0.48 $\pm$ 0.29 \\ 
  Hybrid & Direct C HNR & 0.43 $\pm$ 0.22 & 0.48 $\pm$ 0.18 & 0.32 $\pm$ 0.28 & 0.72 $\pm$ 0.25 & 0.61 $\pm$ 0.42 & 0.39 $\pm$ 0.13 \\ 
  Hybrid & Direct C HWR & 0.43 $\pm$ 0.17 & 0.39 $\pm$ 0.16 & 0.37 $\pm$ 0.23 & 0.85 $\pm$ 0.16 & 0.72 $\pm$ 0.37 & 0.52 $\pm$ 0.15 \\ 
  Hybrid & Composed indirect C HWR & 0.39 $\pm$ 0.24 & 0.44 $\pm$ 0.18 & 0.41 $\pm$ 0.31 & 0.70 $\pm$ 0.31 & 0.68 $\pm$ 0.37 & 0.51 $\pm$ 0.25 \\ 
  Hybrid & Indirect C HWR & 0.39 $\pm$ 0.24 & 0.44 $\pm$ 0.18 & 0.41 $\pm$ 0.31 & 0.70 $\pm$ 0.31 & 0.68 $\pm$ 0.37 & 0.51 $\pm$ 0.25 \\ 
  \midrule
  Single & HNR (1-10 scale) & \cellcolorandbold{Cyan}{0.65 $\pm$ 0.22} & 0.59 $\pm$ 0.16 & 0.60 $\pm$ 0.17 & 0.21 $\pm$ 0.25 & 0.58 $\pm$ 0.2 & 0.25 $\pm$ 0.27 \\ 
  Single & Predict GT from features and HWR & 0.50 $\pm$ 0.18 & 0.47 $\pm$ 0.16 & 0.3 $\pm$ 0.22 & 0.75 $\pm$ 0.19 & 0.57 $\pm$ 0.36 & 0.47 $\pm$ 0.18 \\ 
  Single & Predict GT from features and HNR & 0.47 $\pm$ 0.26 & 0.47 $\pm$ 0.19 & 0.31 $\pm$ 0.25 & 0.75 $\pm$ 0.27 & 0.67 $\pm$ 0.44 & 0.38 $\pm$ 0.11 \\ 
  Single & HWR (1-10 scale) & 0.44 $\pm$ 0.26 & 0.43 $\pm$ 0.22 & 0.71 $\pm$ 0.23 & 0.44 $\pm$ 0.28 & 0.59 $\pm$ 0.24 & 0.52 $\pm$ 0.34 \\ 
  Single & Predict GT from features and C & 0.36 $\pm$ 0.31 & 0.49 $\pm$ 0.2 & 0.3 $\pm$ 0.18 & 0.71 $\pm$ 0.37 & 0.67 $\pm$ 0.37 & 0.35 $\pm$ 0.2 \\ 
  Single & C (1-10 scale) & 0.33 $\pm$ 0.2 & 0.47 $\pm$ 0.14 & \cellcolorandbold{Cyan}{0.21 $\pm$ 0.16} & 0.85 $\pm$ 0.17 & 0.67 $\pm$ 0.41 & 0.50 $\pm$ 0.21 \\ 
  Single & C (binarized $>$=5) & - & 0.39 $\pm$ 0.17 & 0.35 $\pm$ 0.22 & 0.88 $\pm$ 0.25 & 0.88 $\pm$ 0.25 & 0.35 $\pm$ 0.14 \\ 
  Single & HNR (binarized $>$=5) & - & \cellcolorandbold{Cyan}{0.61 $\pm$ 0.17} & 0.65 $\pm$ 0.22 & \cellcolorandbold{Cyan}{0.12 $\pm$ 0.25} & 0.65 $\pm$ 0.14 & \cellcolorandbold{Cyan}{0.12 $\pm$ 0.25} \\ 
  Single & HWR (binarized $>$=5) & - & 0.53 $\pm$ 0.14 & 0.79 $\pm$ 0.16 & 0.15 $\pm$ 0.17 & \cellcolorandbold{Cyan}{0.50 $\pm$ 0.21} & 0.33 $\pm$ 0.41 \\ 
  \midrule
  None & Predict GT from features & 0.48 $\pm$ 0.39 & 0.54 $\pm$ 0.26 & 0.3 $\pm$ 0.2 & 0.62 $\pm$ 0.42 & 0.61 $\pm$ 0.42 & 0.32 $\pm$ 0.23 \\ 
  \midrule
  Random & Randomly pick between C HWR HNR & 0.36 $\pm$ 0.28 & 0.44 $\pm$ 0.2 & 0.53 $\pm$ 0.22 & 0.59 $\pm$ 0.29 & 0.64 $\pm$ 0.25 & 0.46 $\pm$ 0.25 \\ 
  Random & Randomly pick between C HWR & 0.32 $\pm$ 0.2 & 0.34 $\pm$ 0.14 & 0.35 $\pm$ 0.21 & 0.96 $\pm$ 0.09 & 0.86 $\pm$ 0.38 & 0.55 $\pm$ 0.16 \\ 
  Random & Randomly pick between C HNR & 0.3 $\pm$ 0.33 & 0.33 $\pm$ 0.24 & 0.51 $\pm$ 0.32 & 0.83 $\pm$ 0.22 & 0.75 $\pm$ 0.35 & 0.53 $\pm$ 0.22 \\ 
   \bottomrule
\end{tabular}
\end{table*}

\clearpage

\begin{table*}[ht]
\centering
\section*{Appendix D: Result tables for hybrid models for all defendants}
\footnotesize
\caption{Test-set performance of hybrid models built on all individuals, compared to just using a single risk score and other baselines. The benevolent oracle is the risk score best at predicting ground truth, to provide an upper bound on the accuracy reachable on this data set of any hybrid COMPAS-Human model built on the two risk scores. 
The adversarial oracle is the risk score \textbf{worse} at predicting ground truth, to  provide a lower bound. See Table \ref{tab:hybrid_entire_all} for extended version of this table.}
\label{tab:hybrid_entire_all_reduced}
\vspace{0.25cm}
\begin{tabular}{lll}
\toprule
\textbf{Type} & \textbf{Model} & \textbf{AUC} \\
\toprule
\multirow{2}{*}{Oracle} & Benevolent oracle & 0.85 $\pm$ 0.03  \\ 
   & Adversarial oracle & 0.57 $\pm$ 0.03 \\ 

\midrule
  \multirow{3}{*}{Hybrid} & Best hybrid of C and HNR & \cellcolorandbold{Cyan}{0.74 $\pm$ 0.03} \\ 
   & Best hybrid of C and HWR & \cellcolorandbold{Cyan}{0.74 $\pm$ 0.04} \\ 
   & Best hybrid of C, HWR, HNR & 0.73 $\pm$ 0.03  \\ 

 \midrule

   \multirow{6}{*}{Single} & HNR (1-10 scale) & 0.72 $\pm$ 0.03 \\ 
   & HWR (1-10 scale) & 0.72 $\pm$ 0.03 \\ 
   & C (1-10 scale) & 0.71 $\pm$ 0.03 \\ 
   & Predict GT from features and C & 0.71 $\pm$ 0.03 \\
   & Predict GT from features and HWR & 0.71 $\pm$ 0.03\\
   & Predict GT from features and HNR & 0.70$\pm$ 0.03 \\
  \midrule
  None & Predict GT from features & 0.69 $\pm$ 0.02 \\ 
  \midrule
  \multirow{3}{*}{Random} & Randomly pick between C and HWR & 0.73 $\pm$ 0.04 \\ 
   & Randomly pick between C and HNR & 0.72 $\pm$ 0.04 \\ 
   & Randomly pick between C, HWR, HNR & 0.71 $\pm$ 0.03\\ 
 \bottomrule
\end{tabular}
\end{table*}

\begin{table*}[ht]
\centering
\footnotesize
\caption{Test-set performance of hybrid models built on all individuals, compared to just using a single risk score and other baselines. The numbers presented are means and standard deviations calculated over 10 train-test splits.  Best results in cyan and bolded. A reduced version of this table can be seen in Table \ref{tab:hybrid_entire_all_reduced}. Rows marked with $^*$ are the rows labeled as \emph{best} in Table \ref{tab:hybrid_entire_all_reduced}.}
\label{tab:hybrid_entire_all}

\begin{tabular}{llllllll}
\toprule
\textbf{Type} & \textbf{Model} & \textbf{AUC} & \textbf{Bal Acc} & \textbf{FPR} & \textbf{FNR} & \textbf{FDR} & \textbf{FOR} \\
\toprule
Oracle & Benevolent oracle & 0.85 $\pm$ 0.03 & 0.81 $\pm$ 0.02 & 0.19 $\pm$ 0.04 & 0.19 $\pm$ 0.03 & 0.20 $\pm$ 0.04 & 0.18 $\pm$ 0.03 \\ 
  Oracle & Adversarial oracle & 0.57 $\pm$ 0.03 & 0.51 $\pm$ 0.02 & 0.50$\pm$ 0.03 & 0.49 $\pm$ 0.05 & 0.53 $\pm$ 0.03 & 0.46 $\pm$ 0.04 \\ 
\midrule
  Hybrid & Weighted average of C HNR$^*$& \cellcolorandbold{Cyan}{0.74 $\pm$ 0.03} & 0.65 $\pm$ 0.06 & 0.41 $\pm$ 0.21 & 0.29 $\pm$ 0.1 & 0.38 $\pm$ 0.06 & \cellcolorandbold{Cyan}{0.30 $\pm$ 0.04} \\ 
  Hybrid & Weighted average of C HWR$^*$ & \cellcolorandbold{Cyan}{0.74 $\pm$ 0.04} & 0.65 $\pm$ 0.06 & 0.40 $\pm$ 0.2 & 0.29 $\pm$ 0.11 & 0.36 $\pm$ 0.06 & 0.31 $\pm$ 0.05 \\ 
  Hybrid & Direct C HWR HNR$^*$ & 0.73 $\pm$ 0.03 & 0.66 $\pm$ 0.03 & 0.32 $\pm$ 0.05 & 0.36 $\pm$ 0.05 & 0.35 $\pm$ 0.05 & 0.34 $\pm$ 0.05 \\ 
  Hybrid & Direct C HNR & 0.72 $\pm$ 0.04 & 0.65 $\pm$ 0.03 & 0.34 $\pm$ 0.05 & 0.36 $\pm$ 0.06 & 0.38 $\pm$ 0.04 & 0.32 $\pm$ 0.04 \\ 
  Hybrid & Direct C HWR & 0.72 $\pm$ 0.03 & 0.65 $\pm$ 0.03 & 0.32 $\pm$ 0.06 & 0.38 $\pm$ 0.06 & 0.35 $\pm$ 0.06 & 0.35 $\pm$ 0.05 \\ 
 \midrule
  Single & HNR (1-10 scale) & 0.72 $\pm$ 0.03 & 0.66 $\pm$ 0.03 & 0.35 $\pm$ 0.04 & 0.32 $\pm$ 0.04 & 0.37 $\pm$ 0.03 & \cellcolorandbold{Cyan}{0.30 $\pm$ 0.03} \\ 
  Single & HWR (1-10 scale) & 0.72 $\pm$ 0.03 & 0.66 $\pm$ 0.02 & 0.36 $\pm$ 0.04 & 0.31 $\pm$ 0.03 & 0.36 $\pm$ 0.04 & 0.32 $\pm$ 0.04 \\ 
  Single & C (1-10 scale) & 0.71 $\pm$ 0.03 & 0.65 $\pm$ 0.03 & 0.32 $\pm$ 0.03 & 0.38 $\pm$ 0.06 & 0.37 $\pm$ 0.03 & 0.33 $\pm$ 0.04 \\ 
  Single & Predict GT from features and C & 0.71 $\pm$ 0.03 & 0.64 $\pm$ 0.04 & 0.35 $\pm$ 0.04 & 0.36 $\pm$ 0.06 & 0.38 $\pm$ 0.04 & 0.33 $\pm$ 0.05 \\
  Single & Predict GT from features and HWR & 0.71 $\pm$ 0.03 & \cellcolorandbold{Cyan}{0.67 $\pm$ 0.03} & \cellcolorandbold{Cyan}{0.31 $\pm$ 0.05} & 0.36 $\pm$ 0.06 & \cellcolorandbold{Cyan}{0.34 $\pm$ 0.05} & 0.33 $\pm$ 0.06 \\
  Single & Predict GT from features and HNR & 0.70$\pm$ 0.03 & 0.64 $\pm$ 0.02 & 0.35 $\pm$ 0.04 & 0.37 $\pm$ 0.05 & 0.39 $\pm$ 0.03 & 0.33 $\pm$ 0.04 \\
  Single & C (binarized $>$=5) & - & 0.65 $\pm$ 0.03 & 0.32 $\pm$ 0.03 & 0.38 $\pm$ 0.06 & 0.37 $\pm$ 0.03 & 0.33 $\pm$ 0.04 \\ 
  Single & HNR (binarized $>$=5) & - & 0.66 $\pm$ 0.03 & 0.38 $\pm$ 0.04 & 0.30 $\pm$ 0.04 & 0.38 $\pm$ 0.03 & \cellcolorandbold{Cyan}{0.30 $\pm$ 0.04} \\ 
  Single & HWR (binarized $>$=5) & - & 0.66 $\pm$ 0.03 & 0.40 $\pm$ 0.04 & 0.28 $\pm$ 0.04 & 0.37 $\pm$ 0.04 & 0.31 $\pm$ 0.05 \\ 
  \midrule
  None & Predict GT from features & 0.69 $\pm$ 0.02 & 0.63 $\pm$ 0.03 & 0.37 $\pm$ 0.05 & 0.37 $\pm$ 0.06 & 0.40 $\pm$ 0.04 & 0.34 $\pm$ 0.04 \\ 
  \midrule
  Random & Randomly pick between C HWR & 0.73 $\pm$ 0.04 & \cellcolorandbold{Cyan}{0.67 $\pm$ 0.03} & 0.34 $\pm$ 0.03 & 0.32 $\pm$ 0.03 & 0.35 $\pm$ 0.05 & 0.32 $\pm$ 0.04 \\ 
  Random & Randomly pick between C HNR & 0.72 $\pm$ 0.04 & 0.66 $\pm$ 0.04 & 0.33 $\pm$ 0.03 & 0.34 $\pm$ 0.05 & 0.36 $\pm$ 0.04 & 0.31 $\pm$ 0.04 \\ 
  Random & Randomly pick between C HWR HNR & 0.71 $\pm$ 0.03 & \cellcolorandbold{Cyan}{0.67 $\pm$ 0.03} & 0.35 $\pm$ 0.03 & 0.31 $\pm$ 0.05 & 0.37 $\pm$ 0.03 & \cellcolorandbold{Cyan}{0.30 $\pm$ 0.04} \\ 
 \bottomrule
\end{tabular}
\end{table*}

\begin{table*}[ht]
\centering
\footnotesize
\caption{Subgroup (African-Americans) performance of models presented in Table \ref{tab:hybrid_entire_all}.}
\label{tab:hybrid_entire_black}
\begin{tabular}{llllllll}
\toprule
\textbf{Type} & \textbf{Model} & \textbf{AUC} & \textbf{Bal Acc} & \textbf{FPR} & \textbf{FNR} & \textbf{FDR} & \textbf{FOR} \\
\toprule
Oracle & Benevolent oracle & 0.85 $\pm$ 0.03 & 0.81 $\pm$ 0.03 & 0.23 $\pm$ 0.03 & 0.15 $\pm$ 0.04 & 0.18 $\pm$ 0.03 & 0.19 $\pm$ 0.05 \\ 
  Oracle & Adversarial oracle & 0.54 $\pm$ 0.07 & 0.49 $\pm$ 0.05 & 0.60$\pm$ 0.08 & 0.42 $\pm$ 0.04 & 0.46 $\pm$ 0.05 & 0.56 $\pm$ 0.08 \\ 
 \midrule
  Hybrid & Direct C HNR & \cellcolorandbold{Cyan}{0.73 $\pm$ 0.06} & 0.65 $\pm$ 0.05 & 0.47 $\pm$ 0.09 & 0.23 $\pm$ 0.08 & 0.34 $\pm$ 0.05 & 0.34 $\pm$ 0.1 \\ 
  Hybrid & Weighted average of C HNR & \cellcolorandbold{Cyan}{0.73 $\pm$ 0.05} & 0.65 $\pm$ 0.07 & 0.48 $\pm$ 0.18 & 0.22 $\pm$ 0.09 & 0.33 $\pm$ 0.05 & 0.30 $\pm$ 0.14 \\ 
  Hybrid & Direct C HWR HNR & 0.72 $\pm$ 0.03 & 0.65 $\pm$ 0.03 & 0.44 $\pm$ 0.05 & 0.27 $\pm$ 0.06 & 0.30 $\pm$ 0.04 & 0.41 $\pm$ 0.07 \\ 
  Hybrid & Weighted average of C HWR & 0.72 $\pm$ 0.04 & 0.64 $\pm$ 0.06 & 0.47 $\pm$ 0.19 & 0.25 $\pm$ 0.1 & 0.30 $\pm$ 0.05 & 0.41 $\pm$ 0.07 \\ 
  Hybrid & Weighted average of C HWR HNR & 0.71 $\pm$ 0.05 & 0.62 $\pm$ 0.07 & 0.59 $\pm$ 0.24 & \cellcolorandbold{Cyan}{0.17 $\pm$ 0.13} & 0.36 $\pm$ 0.07 & \cellcolorandbold{Cyan}{0.23 $\pm$ 0.18} \\ 
  Hybrid & Direct C HWR & 0.70$\pm$ 0.04 & 0.62 $\pm$ 0.02 & 0.47 $\pm$ 0.07 & 0.29 $\pm$ 0.06 & 0.32 $\pm$ 0.05 & 0.43 $\pm$ 0.05 \\ 
  \midrule
  Single & Predict GT from features and HNR & 0.71 $\pm$ 0.05 & 0.63 $\pm$ 0.04 & 0.49 $\pm$ 0.08 & 0.24 $\pm$ 0.08 & 0.35 $\pm$ 0.05 & 0.36 $\pm$ 0.1 \\ 
  Single & HNR (1-10 scale) & 0.71 $\pm$ 0.04 & \cellcolorandbold{Cyan}{0.68 $\pm$ 0.04} & \cellcolorandbold{Cyan}{0.39 $\pm$ 0.06} & 0.26 $\pm$ 0.05 & 0.30 $\pm$ 0.04 & 0.34 $\pm$ 0.07 \\ 
  Single & Predict GT from features and C & 0.70$\pm$ 0.04 & 0.62 $\pm$ 0.04 & 0.49 $\pm$ 0.08 & 0.27 $\pm$ 0.06 & 0.32 $\pm$ 0.05 & 0.43 $\pm$ 0.08 \\ 
  Single & HWR (1-10 scale) & 0.70$\pm$ 0.03 & 0.65 $\pm$ 0.03 & 0.43 $\pm$ 0.05 & 0.27 $\pm$ 0.04 & \cellcolorandbold{Cyan}{0.29 $\pm$ 0.04} & 0.41 $\pm$ 0.05 \\ 
  Single & C (1-10 scale) & 0.69 $\pm$ 0.05 & 0.63 $\pm$ 0.04 & 0.43 $\pm$ 0.06 & 0.31 $\pm$ 0.07 & 0.34 $\pm$ 0.05 & 0.39 $\pm$ 0.07 \\ 
  Single & Predict GT from features and HWR & 0.69 $\pm$ 0.04 & 0.64 $\pm$ 0.04 & 0.45 $\pm$ 0.07 & 0.26 $\pm$ 0.06 & 0.30 $\pm$ 0.05 & 0.40 $\pm$ 0.07 \\ 
  Single & C (binarized $>$=5) & - & 0.63 $\pm$ 0.04 & 0.43 $\pm$ 0.06 & 0.31 $\pm$ 0.07 & 0.34 $\pm$ 0.05 & 0.39 $\pm$ 0.07 \\ 
  Single & HNR (binarized $>$=5) & - & 0.67 $\pm$ 0.04 & 0.41 $\pm$ 0.07 & 0.25 $\pm$ 0.05 & 0.32 $\pm$ 0.05 & 0.34 $\pm$ 0.07 \\ 
  Single & HWR (binarized $>$=5) & - & 0.64 $\pm$ 0.03 & 0.48 $\pm$ 0.05 & 0.24 $\pm$ 0.05 & 0.31 $\pm$ 0.04 & 0.39 $\pm$ 0.07 \\ 
  \midrule
  None & Predict GT from features & 0.69 $\pm$ 0.04 & 0.62 $\pm$ 0.04 & 0.52 $\pm$ 0.06 & 0.23 $\pm$ 0.08 & 0.36 $\pm$ 0.05 & 0.36 $\pm$ 0.11 \\ 
  \midrule
  Random & Randomly pick between C HWR & 0.72 $\pm$ 0.04 & 0.66 $\pm$ 0.03 & 0.44 $\pm$ 0.04 & 0.24 $\pm$ 0.04 & \cellcolorandbold{Cyan}{0.29 $\pm$ 0.04} & 0.38 $\pm$ 0.05 \\ 
  Random & Randomly pick between C HNR & 0.70$\pm$ 0.04 & 0.65 $\pm$ 0.04 & 0.42 $\pm$ 0.05 & 0.27 $\pm$ 0.05 & 0.32 $\pm$ 0.04 & 0.36 $\pm$ 0.07 \\ 
  Random & Randomly pick between C HWR HNR & 0.70$\pm$ 0.05 & 0.66 $\pm$ 0.04 & 0.42 $\pm$ 0.07 & 0.25 $\pm$ 0.05 & 0.32 $\pm$ 0.05 & 0.35 $\pm$ 0.08 \\ 
 \bottomrule
\end{tabular}
\end{table*}

\begin{table*}[ht]
\centering
\footnotesize
\caption{Subgroup (whites) performance of models presented in Table
\ref{tab:hybrid_entire_all}.}
\label{tab:hybrid_entire_white}
\begin{tabular}{llllllll}
\toprule
\textbf{Type} & \textbf{Model} & \textbf{AUC} & \textbf{Bal Acc} & \textbf{FPR} & \textbf{FNR} & \textbf{FDR} & \textbf{FOR} \\
\toprule
Oracle & Benevolent oracle & 0.84 $\pm$ 0.04 & 0.8 $\pm$ 0.03 & 0.13 $\pm$ 0.06 & 0.28 $\pm$ 0.04 & 0.23 $\pm$ 0.07 & 0.16 $\pm$ 0.05 \\ 
  Oracle & Adversarial oracle & 0.55 $\pm$ 0.05 & 0.48 $\pm$ 0.05 & 0.42 $\pm$ 0.06 & 0.61 $\pm$ 0.09 & 0.63 $\pm$ 0.08 & 0.4 $\pm$ 0.07 \\ 

\midrule
  Hybrid & Weighted average of C HWR & \cellcolorandbold{Cyan}{0.75 $\pm$ 0.05} & 0.65 $\pm$ 0.06 & 0.34 $\pm$ 0.22 & 0.37 $\pm$ 0.14 & 0.46 $\pm$ 0.1 & \cellcolorandbold{Cyan}{0.23 $\pm$ 0.1} \\ 
  Hybrid & Weighted average of C HWR HNR & 0.74 $\pm$ 0.05 & 0.57 $\pm$ 0.08 & 0.64 $\pm$ 0.36 & \cellcolorandbold{Cyan}{0.21 $\pm$ 0.22} & 0.55 $\pm$ 0.09 & \cellcolorandbold{Cyan}{0.23 $\pm$ 0.31} \\ 
  Hybrid & Weighted average of C HNR & 0.72 $\pm$ 0.03 & 0.62 $\pm$ 0.07 & 0.35 $\pm$ 0.23 & 0.4 $\pm$ 0.16 & 0.46 $\pm$ 0.1 & 0.26 $\pm$ 0.11 \\ 
  Hybrid & Direct C HWR & 0.70$\pm$ 0.04 & 0.62 $\pm$ 0.05 & 0.21 $\pm$ 0.07 & 0.54 $\pm$ 0.1 & 0.43 $\pm$ 0.12 & 0.29 $\pm$ 0.06 \\ 
  Hybrid & Direct C HWR HNR & 0.70$\pm$ 0.04 & 0.63 $\pm$ 0.04 & 0.21 $\pm$ 0.05 & 0.53 $\pm$ 0.08 & 0.44 $\pm$ 0.1 & 0.29 $\pm$ 0.06 \\ 
  Hybrid & Direct C HNR & 0.67 $\pm$ 0.05 & 0.62 $\pm$ 0.04 & 0.22 $\pm$ 0.04 & 0.55 $\pm$ 0.07 & 0.43 $\pm$ 0.1 & 0.31 $\pm$ 0.05 \\ 
  \midrule
  Single & HWR (1-10 scale) & 0.74 $\pm$ 0.05 & \cellcolorandbold{Cyan}{0.67 $\pm$ 0.05} & 0.29 $\pm$ 0.07 & 0.38 $\pm$ 0.06 & 0.44 $\pm$ 0.08 & 0.24 $\pm$ 0.07 \\ 
  Single & HNR (1-10 scale) & 0.70$\pm$ 0.04 & 0.62 $\pm$ 0.05 & 0.33 $\pm$ 0.04 & 0.43 $\pm$ 0.06 & 0.47 $\pm$ 0.05 & 0.29 $\pm$ 0.07 \\ 
  Single & C (1-10 scale) & 0.69 $\pm$ 0.06 & 0.63 $\pm$ 0.05 & 0.24 $\pm$ 0.05 & 0.50$\pm$ 0.12 & 0.44 $\pm$ 0.1 & 0.29 $\pm$ 0.05 \\ 
  Single & Predict GT from features and HWR & 0.69 $\pm$ 0.05 & 0.63 $\pm$ 0.06 & 0.2 $\pm$ 0.06 & 0.54 $\pm$ 0.09 & 0.43 $\pm$ 0.12 & 0.29 $\pm$ 0.06 \\ 
  Single & Predict GT from features and C & 0.67 $\pm$ 0.05 & 0.63 $\pm$ 0.03 & \cellcolorandbold{Cyan}{0.19 $\pm$ 0.05} & 0.55 $\pm$ 0.05 & \cellcolorandbold{Cyan}{0.4 $\pm$ 0.09} & 0.3 $\pm$ 0.05 \\ 
  Single & Predict GT from features and HNR & 0.66 $\pm$ 0.06 & 0.61 $\pm$ 0.05 & 0.23 $\pm$ 0.07 & 0.56 $\pm$ 0.08 & 0.45 $\pm$ 0.11 & 0.32 $\pm$ 0.05 \\ 
  Single & C (binarized $>$=5) & - & 0.63 $\pm$ 0.05 & 0.24 $\pm$ 0.05 & 0.50$\pm$ 0.12 & 0.44 $\pm$ 0.1 & 0.29 $\pm$ 0.05 \\ 
  Single & HNR (binarized $>$=5) & - & 0.63 $\pm$ 0.05 & 0.36 $\pm$ 0.04 & 0.38 $\pm$ 0.06 & 0.47 $\pm$ 0.05 & 0.27 $\pm$ 0.07 \\ 
  Single & HWR (binarized $>$=5) & - & 0.66 $\pm$ 0.04 & 0.31 $\pm$ 0.07 & 0.36 $\pm$ 0.05 & 0.45 $\pm$ 0.07 & 0.24 $\pm$ 0.07 \\ 
  \midrule
  None & Predict GT from features & 0.65 $\pm$ 0.05 & 0.60$\pm$ 0.05 & 0.22 $\pm$ 0.08 & 0.57 $\pm$ 0.08 & 0.44 $\pm$ 0.13 & 0.32 $\pm$ 0.05 \\ 
  \midrule
  Random & Randomly pick between C HWR & 0.73 $\pm$ 0.04 & 0.64 $\pm$ 0.03 & 0.26 $\pm$ 0.07 & 0.46 $\pm$ 0.05 & 0.45 $\pm$ 0.08 & 0.27 $\pm$ 0.06 \\ 
  Random & Randomly pick between C HNR & 0.72 $\pm$ 0.06 & 0.65 $\pm$ 0.06 & 0.25 $\pm$ 0.03 & 0.45 $\pm$ 0.09 & 0.42 $\pm$ 0.07 & 0.28 $\pm$ 0.06 \\ 
  Random & Randomly pick between C HWR HNR & 0.70$\pm$ 0.03 & 0.65 $\pm$ 0.05 & 0.3 $\pm$ 0.05 & 0.4 $\pm$ 0.08 & 0.44 $\pm$ 0.07 & 0.27 $\pm$ 0.06 \\ 
\bottomrule
\end{tabular}
\end{table*}

\begin{table*}[ht]
\centering
\footnotesize
\caption{Subgroup (other races) performance of models presented in Table \ref{tab:hybrid_entire_all}.}
\label{tab:hybrid_entire_other}
\begin{tabular}{llllllll}
\toprule
\textbf{Type} & \textbf{Model} & \textbf{AUC} & \textbf{Bal Acc} & \textbf{FPR} & \textbf{FNR} & \textbf{FDR} & \textbf{FOR} \\
 \toprule
Oracle & Benevolent oracle & 0.84 $\pm$ 0.12 & 0.82 $\pm$ 0.09 & 0.09 $\pm$ 0.1 & 0.27 $\pm$ 0.17 & 0.16 $\pm$ 0.17 & 0.14 $\pm$ 0.12 \\ 
  Oracle & Adversarial oracle & 0.48 $\pm$ 0.1 & 0.46 $\pm$ 0.1 & 0.45 $\pm$ 0.14 & 0.62 $\pm$ 0.17 & 0.70$\pm$ 0.1 & 0.38 $\pm$ 0.17 \\ 
 \midrule
  Hybrid & Weighted average of C HWR HNR & \cellcolorandbold{Cyan}{0.76 $\pm$ 0.15} & \cellcolorandbold{Cyan}{0.74 $\pm$ 0.18} & \cellcolorandbold{Cyan}{0.18 $\pm$ 0.12} & 0.33 $\pm$ 0.37 & \cellcolorandbold{Cyan}{0.41 $\pm$ 0.32} & \cellcolorandbold{Cyan}{0.14 $\pm$ 0.13} \\ 
  Hybrid & Weighted average of C HNR & 0.75 $\pm$ 0.15 & 0.68 $\pm$ 0.17 & 0.30 $\pm$ 0.27 & 0.34 $\pm$ 0.29 & 0.44 $\pm$ 0.27 & 0.25 $\pm$ 0.28 \\ 
  Hybrid & Direct C HNR & 0.69 $\pm$ 0.12 & 0.56 $\pm$ 0.17 & 0.30 $\pm$ 0.1 & 0.58 $\pm$ 0.28 & 0.62 $\pm$ 0.24 & 0.30 $\pm$ 0.16 \\ 
  Hybrid & Direct C HWR & 0.69 $\pm$ 0.15 & 0.59 $\pm$ 0.12 & 0.21 $\pm$ 0.14 & 0.61 $\pm$ 0.24 & 0.55 $\pm$ 0.3 & 0.29 $\pm$ 0.14 \\ 
  Hybrid & Direct C HWR HNR & 0.68 $\pm$ 0.08 & 0.58 $\pm$ 0.08 & 0.27 $\pm$ 0.09 & 0.58 $\pm$ 0.2 & 0.58 $\pm$ 0.21 & 0.28 $\pm$ 0.15 \\ 
  Hybrid & Weighted average of C HWR & 0.66 $\pm$ 0.07 & 0.62 $\pm$ 0.07 & 0.31 $\pm$ 0.12 & 0.44 $\pm$ 0.13 & 0.52 $\pm$ 0.17 & 0.25 $\pm$ 0.11 \\ 
    \midrule
  Single & HNR (1-10 scale) & 0.73 $\pm$ 0.16 & 0.69 $\pm$ 0.14 & 0.32 $\pm$ 0.17 & 0.31 $\pm$ 0.2 & 0.44 $\pm$ 0.19 & 0.21 $\pm$ 0.18 \\ 
  Single & Predict GT from features and C & 0.69 $\pm$ 0.12 & 0.56 $\pm$ 0.17 & 0.29 $\pm$ 0.08 & 0.59 $\pm$ 0.27 & 0.60 $\pm$ 0.21 & 0.30 $\pm$ 0.15 \\ 
  Single & Predict GT from features and HNR & 0.69 $\pm$ 0.14 & 0.60$\pm$ 0.21 & 0.26 $\pm$ 0.13 & 0.54 $\pm$ 0.37 & 0.55 $\pm$ 0.3 & 0.29 $\pm$ 0.19 \\ 
  Single & Predict GT from features and HWR & 0.67 $\pm$ 0.14 & 0.65 $\pm$ 0.1 & 0.22 $\pm$ 0.12 & 0.48 $\pm$ 0.17 & 0.45 $\pm$ 0.19 & 0.25 $\pm$ 0.13 \\
  Single & HWR (1-10 scale) & 0.66 $\pm$ 0.07 & 0.61 $\pm$ 0.05 & 0.37 $\pm$ 0.1 & 0.41 $\pm$ 0.12 & 0.54 $\pm$ 0.14 & 0.26 $\pm$ 0.1 \\ 
  Single & C (1-10 scale) & 0.64 $\pm$ 0.11 & 0.61 $\pm$ 0.09 & 0.20 $\pm$ 0.1 & 0.57 $\pm$ 0.14 & 0.46 $\pm$ 0.13 & 0.28 $\pm$ 0.14 \\ 
  Single & C (binarized $>$=5) & - & 0.61 $\pm$ 0.09 & 0.20 $\pm$ 0.1 & 0.57 $\pm$ 0.14 & 0.46 $\pm$ 0.13 & 0.28 $\pm$ 0.14 \\ 
  Single & HNR (binarized $>$=5) & - & 0.71 $\pm$ 0.14 & 0.34 $\pm$ 0.17 & \cellcolorandbold{Cyan}{0.24 $\pm$ 0.2} & 0.43 $\pm$ 0.19 & 0.18 $\pm$ 0.17 \\ 
  Single & HWR (binarized $>$=5) & - & 0.64 $\pm$ 0.08 & 0.43 $\pm$ 0.11 & 0.30 $\pm$ 0.14 & 0.54 $\pm$ 0.11 & 0.23 $\pm$ 0.12 \\ 
  \midrule
  None & Predict GT from features & 0.68 $\pm$ 0.16 & 0.57 $\pm$ 0.21 & 0.32 $\pm$ 0.14 & 0.53 $\pm$ 0.35 & 0.59 $\pm$ 0.28 & 0.30 $\pm$ 0.19 \\
  \midrule
  Random & Randomly pick between C HWR HNR & 0.66 $\pm$ 0.13 & 0.60$\pm$ 0.09 & 0.30 $\pm$ 0.11 & 0.51 $\pm$ 0.14 & 0.53 $\pm$ 0.14 & 0.28 $\pm$ 0.15 \\ 
  Random & Randomly pick between C HNR & 0.62 $\pm$ 0.24 & 0.63 $\pm$ 0.18 & 0.26 $\pm$ 0.17 & 0.48 $\pm$ 0.28 & 0.48 $\pm$ 0.28 & 0.26 $\pm$ 0.18 \\ 
  Random & Randomly pick between C HWR & 0.58 $\pm$ 0.14 & 0.55 $\pm$ 0.08 & 0.28 $\pm$ 0.12 & 0.61 $\pm$ 0.15 & 0.57 $\pm$ 0.12 & 0.32 $\pm$ 0.14 \\ 
  \bottomrule
\end{tabular}
\end{table*}

\end{document}